\documentclass[conference]{IEEEtran}
\usepackage{times}

\usepackage[numbers]{natbib}
\usepackage{multicol}
\usepackage[bookmarks=true]{hyperref}
\usepackage{graphicx}
\usepackage[font=small,labelfont=bf,justification=justified]{caption}
\usepackage{subcaption}
\usepackage{amsmath}
\usepackage{amssymb}
\usepackage{gensymb}
\usepackage{bm}
\usepackage{float}
\usepackage{color}
\usepackage{multirow}
\usepackage{hhline}
\usepackage{changepage}
\usepackage{comment}

\usepackage[ruled,vlined,linesnumbered]{algorithm2e}

\newcommand{\nonl}{\renewcommand{\nl}{\let\nl\oldnl}}

\begin{document}

\title{Human-Guided Planning for Complex Manipulation Tasks Using the Screw Geometry of Motion}

% You will get a Paper-ID when submitting a pdf file to the conference system
%\author{Author Names Omitted for Anonymous Review. Paper-ID [add your ID here]}

\author{\authorblockN{Dasharadhan Mahalingam}
\authorblockA{Department of Mechanical Engineering\\
Stony Brook University\\
Stony Brook, NY 11794\\
Email: dasharadhan.mahalingam@stonybrook.edu}
\and
\authorblockN{Nilanjan Chakraborty}
\authorblockA{ Department of Mechanical Engineering\\
Stony Brook University\\
Stony Brook, NY 11794\\
Email: nilanjan.chakraborty@stonybrook.edu}}
%\and
%\authorblockN{James Kirk\\ and Montgomery Scott}
%\authorblockA{Starfleet Academy\\
%San Francisco, California 96678-2391\\

% avoiding spaces at the end of the author lines is not a problem with
% conference papers because we don't use \thanks or \IEEEmembership

%Homer Simpson\authorrefmark{2},
%Montgomery Scott\authorrefmark{3} and
%Eldon Tyrell\authorrefmark{4}}
%Telephone: (800) 555--1212, Fax: (888) 555--1212}
%\authorblockA{\authorrefmark{4}Tyrell Inc., 123 Replicant Street, Los Angeles, California 90210--4321}}

\maketitle

\begin{abstract}
In this paper, we present a novel method of motion planning for performing complex manipulation tasks by using human demonstration and exploiting the screw geometry of motion. We consider complex manipulation tasks where there are constraints on the motion of the end effector of the robot. Examples of such tasks include opening a door, opening a drawer, transferring granular material from one container to another with a spoon, and loading dishes to a dishwasher. Our approach consists of two steps: First, using the fact that a motion in the task space of the robot can be approximated by using a sequence of constant screw motions, we segment a human demonstration into a sequence of constant screw motions. Second, we use the segmented screws to generate motion plans via screw-linear interpolation for other instances of the same task. The use of screw segmentation allows us to capture the invariants of the demonstrations in a coordinate-free fashion, thus allowing us to plan for different task instances from just one example. We present extensive experimental results on a variety of manipulation scenarios showing that our method can be used across a wide range of manipulation tasks.

%In this paper, we present a \textit{transductive learning approach} to learn from demonstration for robotic manipulation tasks. We use a novel combination of kinesthetic demonstration along with geometric screw theory of rigid body motion interpolation to develop a transductive learning approach. We show that, using our approach, just from one demonstration we are able to generalize from one task instance to another for fairly complex tasks like scooping and moving contents from one bowl to another and loading a dish in a dishwasher rack. 
\end{abstract}

\IEEEpeerreviewmaketitle

\section{Introduction}
\begin{figure}[t!]
    \begin{subfigure}[b]{0.24\textwidth}
        \includegraphics[width=\textwidth]{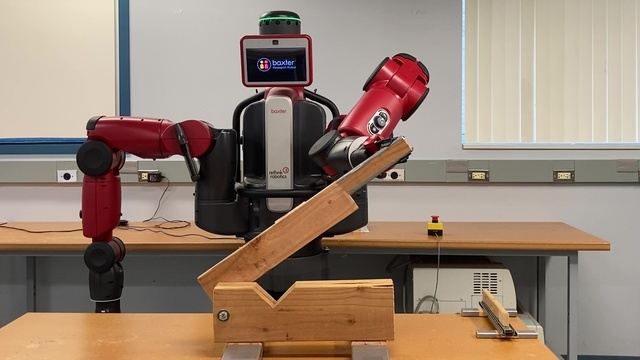}
    \end{subfigure}
    \begin{subfigure}[b]{0.24\textwidth}
        \includegraphics[width=\textwidth]{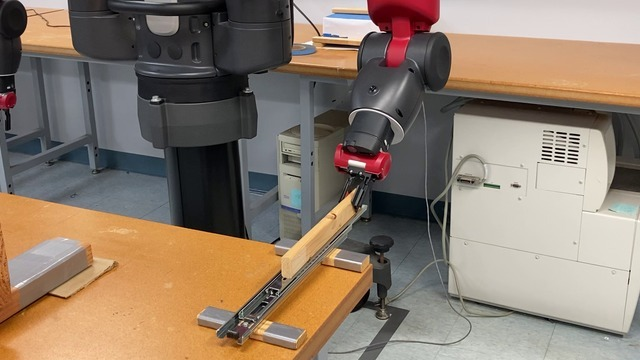}
    \end{subfigure}
    \par\smallskip
    \begin{subfigure}[b]{0.24\textwidth}
        \includegraphics[width=\textwidth]{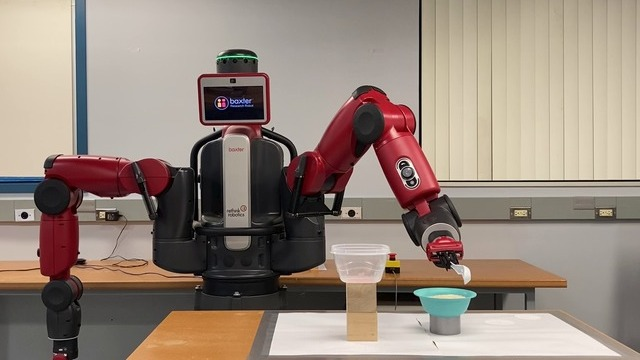}
    \end{subfigure}
    \begin{subfigure}[b]{0.24\textwidth}
        \includegraphics[width=\textwidth]{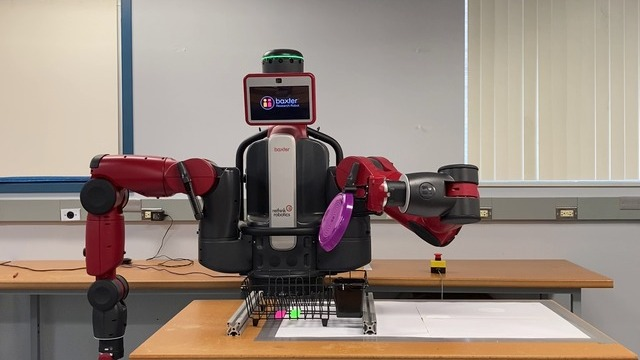}
    \end{subfigure}
    \caption{\textbf{\textsc{Typical Complex Manipulation Tasks:}} (Clockwise from top left) Manipulation of object constrained by a revolute joint; Manipulation of object constrained by a prismatic joint; Arranging dishes in a dish rack; Scooping and pouring}
    \label{fig:task_examples}
\end{figure}

The use of robot manipulators in service industry, domestic environments, as well as in assistive robotics to help people who have lost the use of hands, depends on the ability of the robots to perform {\em complex manipulation tasks}. Complex manipulation tasks, as opposed to simple reaching or pick-and-place tasks, are characterized by the presence of constraints on the motion of the end effector of the robot. Figure \ref{fig:task_examples} shows some typical complex manipulation tasks, where the constraints can arise from the mechanical structure of the objects (top row) or from the nature of the task itself (bottom row). Operating articulated objects (like opening/closing  doors, windows, drawers, and bottle caps) imposes constraints on the motion of the robot end effector depending on the type of the joint (like revolute, prismatic, and helical). 

Furthermore, tasks like scooping and pouring, loading a dish into a dishwasher rack, requires constraints on the motion of the end effector. These constraints essentially characterize the task (please see the results section for a more detailed discussion) and the constraints may change during the motion. For example, in the scooping and pouring task in Figure~\ref{fig:task_examples}, the spoon should not rotate while transferring, it should not translate while pouring and during scooping there is a different constraint on the motion which is hard to describe. 

Typically, for the robot to perform the task, a robotics expert needs to come up with a mathematical representation of these constraints (which may not always be easy, think about the scooping task) and they need to be pre-programmed. However, many modern light weight cobots like Baxter (Rethink Robotics), Panda (Franka Emika) and UR-3, UR-5, UR-10 (Universal Robots) are designed with easily accessible tools to allow a non-expert in robotics to give a kinesthetic demonstration, i.e., hold the hand of the robot and show how to do a task. Any constraint that characterizes the task is {\em embedded in these demonstrations}. Therefore, in principle, it is possible to utilize these constraints (even from a single example) for motion planning for a different instance of the task. Note that an instance of a task, or task instance for short, is defined by the poses of task-related objects. Different task instances correspond to different poses of the task-related objects for the same task.
{\em The goal of this paper is to develop a method that can utilize the embedded task constraints in a single kinesthetic demonstration of a task from a human to plan motions for a different instance of the same task}.

%One class of complex manipulation tasks include operating articulated objects whose motion is constrained by mechanical joints like revolute or prismatic or helical joints. Such objects are omnipresent in semi-structured environments and products designed for humans. Examples include, room doors, cabinet doors, windows, drawers, bottle caps, etc. By the nature of their design, operating such objects, e.g., opening a door or a drawer, imposes constraints on the motion of the end effector of the robot, while performing the manipulation. More general complex manipulation tasks like pouring liquid, scooping granular material (e.g., scooping cereal with spoon to feed a person), loading a dishwasher, etc., also requires constraints on the motion of the end effector during the manipulation task and these constraints essentially characterize the task. 
%\textcolor{red}{Put in a figure here showing the four tasks that we are considering}.
%Therefore, the goal of this paper is to develop motion planning algorithms for complex manipulation tasks. 

The end-effector motion of a robot is a curve in the group of rigid body motions, i.e., $SE(3)$. Our approach is based on exploiting the following characteristic of rigid body motion in $SE(3)$ which is implied by Chasles theorem: {\em Any curve in $SE(3)$, which represents a rigid body motion, can be approximated arbitrarily closely by a sequence of constant screw motions (or one-parameter subgroups of $SE(3)$)}. Note that when the end-effector motion is constrained by a mechanical joint, its motion is a single constant screw motion, i.e., the motion is constrained to lie in a one-parameter subgroup of $SE(3)$~\cite{sclerp_motion_planner}. 

Based on the above observation, we present a novel two-step solution approach: (a) Given a kinesthetic demonstration of a task, we segment the motion of the end effector in the task space into a sequence of constant screws.  (b) For the new task instance, using the segmented screws, we compute a motion plan based on Screw linear interpolation (ScLERP) which automatically ensures that the constant screw constraints embedded in the demonstrated motion are satisfied. The sequence of constant screws represent the task constraints in a coordinate invariant manner, i.e., it does not depend on the location of the reference frame on the end-effector. By extracting the task constraints which are enforced either due to the presence of joints or due to the inherent nature of the task, we are able to generalize a single demonstration to different task instances. This two-step motion planning method is the {\em key contribution} of this paper. We also provide extensive experimental results that validates the ability of our approach to generalize from a single example. The key novelty of our method is the combination of data with classical ideas in screw theory that exploits the natural kinematic structure of motion. We show that exploiting the structure of motion can lead to data-efficient approaches for planning using human demonstration as a guide.

\section{Related Work}
\label{sec:rw}

The user-guided manipulation task planning approach presented in this paper is based on the hypothesis that, in principle, even one successful (kinesthetic) demonstration can be used to either implicitly or explicitly encode the constraints that characterize the task and generate motion plans for new instances of a task. Coordinate free shape descriptors of rigid body motion that compactly describe a given demonstration on an object and transfer to new instances with new initial and goal pose of the object is proposed in~\cite{DeSchutter2009,Vochten2019}. However, the constraints present in the tasks are not explicitly extracted (i.e., the motion invariants are not the constraints). Although these methods~\cite{DeSchutter2009,Vochten2019} can be used for tasks like pouring where there is a single constraint, for more complex tasks like scooping and pouring (that we consider in this paper) where the task constraints change during the motion, it is not clear that one can use these methods directly.

Another approach to generate motions from a single demonstration is the use of Dynamical movement primitives (DMPs)~\cite{IjspeertNHPS13, HerschGCB08, pastor2009, Saveriano2019}.  While this is an elegant bio-inspired approach, DMPs again do not consider task constraints explicitly. Furthermore, they are not coordinate-invariant, i.e., their performance depend on the coordinate system attached to the end-effector, which does not allow them to generalize to regions of the task space that are not near the demonstration~\cite{Vochten2019}. In contrast, our method explicitly extracts the task constraints which are expressed as a sequence of constant screw motions that are coordinate-invariant.

Probabilistic approaches to Learning from Demonstration~\cite{Billard2016}, e.g., those using Gaussian Mixture Models (GMM) \cite{calinon2007} or Hidden Markov Models (HMM) \cite{calinon2010, calinon2011} require multiple demonstrations.
Trajectory segmentation based approaches that decompose complex movements as a sequence of primitive motions such as \cite{schaal2011, Lee2015} either require a predefined library of primitive motions or multiple demonstrations of a task to extract the sequence of primitive motions.
Approaches to manipulate articulated objects such as \cite{sturm2009, sturm2011} try to estimate the kinematic model (prismatic/revolute/rigid) which maximizes the posterior probability of the object pose observations. 
In~\cite{jainscrewnet20}, the authors use a neural network to predict the screw parameters of articulated objects from observed depth images. However, these methods cannot be extended to manipulation of non-articulated objects where the motion constraints arise from the nature of the tasks, and some of them are also data hungry.
%Approaches to manipulate articulated objects such as \cite{jainscrewnet20}, \cite{sturm2009}, \cite{sturm2011}, \cite{abbatematteo2020} determine the joint constraints explicitly from provided demonstrations or observations.
%This would allow to compute the motion plan required for a new task instance by applying this geometric constraint to the motion of the end-effector.
%While the LfD approaches mentioned above can be applied to either: 1) Manipulation of physically unconstrained objects or 2) Manipulation of objects constrained by physical  joints, they might fail or not generalize well on tasks belonging to the other category.
%Our approach aims to utilize a unified approach that works well for tasks belonging to both categories.
In contrast, our approach aims to extract the motion constraints of both articulated and non-articulated object using a single demonstration.

%A task space based approach to use provided demonstrations for performing tasks by interpolating between guiding poses using screw linear interpolation has been studied previously in \cite{user_guided_planning}. 
This work builds on our prior work in~\cite{user_guided_planning, laha2022}. The key distinction again is that here we extract the task constraints explicitly as a sequence of constant screw motions, which allows us to perform a more complex set of tasks.  
%for manipulation of constrained objects and to also provide a better heuristic for selecting guiding poses from a demonstration. We segment the motion as a sequence of constant screws and enforce this sequence of constant screw constraints on the end-effector. Motion Planning with screw motion constraints on the end-effector has been explored in \cite{sclerp_motion_planner}. By extracting the sequence of constant screw motion constraints that are enforced on the end-effector for everyday tasks from provided demonstrations, the required motion for a new task instance is determined using \cite{sclerp_motion_planner} which ensures that the constant screw constraints are satisfied.

\section{Mathematical Preliminaries}
\label{sec:prelim}
In this section, we present the background knowledge on rigid body motion and screw geometry of rigid body motion required for this work.

\noindent
\textbf{Quaternions and Rotations}: The quaternions are the set of hypercomplex numbers, $\mathbb{H}$. A quaternion $\mathbf{Q} \in \mathbb{H}$ can be represented as a 4-tuple $\mathbf{Q} = (q_0, \boldsymbol{q}_r) = (q_0, q_1, q_2, q_3)$, $q_0 \in \mathbb{R}$ is the real scalar part, 
$\boldsymbol{q}_r=(q_1, q_2, q_3) \in \mathbb{R}^3$ corresponds to the imaginary part.
The conjugate, norm, and inverse of a quaternion $\mathbf{Q}$ is given by
%$\mathbf{Q}^* = q_0 - \boldsymbol{q}$
$\mathbf{Q}^* = (q_0, -\boldsymbol{q}_r)$, $\lVert \mathbf{Q} \rVert = \sqrt{\mathbf{Q} \mathbf{Q}^*} = \sqrt{\mathbf{Q}^* \mathbf{Q}}$,
%  = \sqrt{q_0^2 + q_1^2 + q_2^2 + q_3^2}
and $\mathbf{Q}^{-1} = \mathbf{Q}^*/{\lVert \mathbf{Q} \rVert}^2$, respectively. Addition and multiplication of two quaternions
% $\mathbf{P} = p_0 + \boldsymbol{p} =(p_0, \boldsymbol{p}_r)$
% $\mathbf{Q} = q_0 + \boldsymbol{q} = (q_0, \boldsymbol{q}_r)$
$\mathbf{P} = (p_0, \boldsymbol{p}_r)$ and
$\mathbf{Q} = (q_0, \boldsymbol{q}_r)$ are performed as $\mathbf{P}+\mathbf{Q} = (p_0 + q_0, \boldsymbol{p}_r + \boldsymbol{q}_r)$ and $\mathbf{P}\mathbf{Q} = (p_0 q_0 - \boldsymbol{p}_r \cdot \boldsymbol{q}_r, p_0 \boldsymbol{q}_r + q_0 \boldsymbol{p}_r + \boldsymbol{p}_r \times \boldsymbol{q}_r)$.
% \orange{As a result, $\mathbf{P}\mathbf{Q} \ne \mathbf{Q}\mathbf{P}$, $(\mathbf{P}\mathbf{Q})^* = \mathbf{Q}^* \mathbf{P}^*$, $\lVert \mathbf{P}\mathbf{Q} \rVert = \lVert \mathbf{P} \rVert \lVert \mathbf{Q} \rVert$, and $\mathbf{Q}\mathbf{Q}^{-1}=\mathbf{Q}^{-1}\mathbf{Q}=1$.}
The quaternion $\mathbf{Q}$ is a \textit{unit quaternion}
% or \textit{versor}
if ${\lVert \mathbf{Q} \rVert} = 1$, and consequently, $\mathbf{Q}^{-1} = \mathbf{Q}^*$. Unit quaternions are used to represent the set of all rigid body rotations,  $SO(3)$, the Special Orthogonal group of dimension $3$. Mathematically,  $SO(3)=\left\{\boldsymbol{R} \in \mathbb{R}^{3 \times 3}\left|\boldsymbol{R}^{\mathrm{T}} \boldsymbol{R}=\boldsymbol{R} \boldsymbol{R}^{\boldsymbol{T}}=\boldsymbol{I}_3,\right| \boldsymbol{R} \mid=1\right\}$, where $\boldsymbol{I}_3$ is a $3\times3$ identity matrix and $\left| \cdot \right|$ is the determinant operator. The unit quaternion corresponding to a rotation is $\mathbf{Q}_R = (\cos\frac{\theta}{2}, \boldsymbol{\omega} \sin\frac{\theta}{2})$, where $\theta \in [0,\pi]$ is the angle of rotation about a unit axis $\boldsymbol{\omega} \in \mathbb{R}^3$. 

\noindent
\textbf{Dual Quaternions and Rigid Displacements}:
In general, dual numbers are defined as $d = a + \epsilon b$ where $a$ and $b$ are elements of an algebraic field, and $\epsilon$ is a \textit{dual unit} with $\epsilon ^ 2 = 0, \epsilon \ne 0$.
%\orange{Moreover, a function $f(a + \epsilon b)$ can be expanded using the Taylor expansion as $f(a)+\epsilon bf'(a)$.}
Similarly, a dual quaternion $\mathbf{D}$ is defined as $\mathbf{D}= \mathbf{P} + \epsilon \mathbf{Q}$
% or an 8-tuple $\mathbf{D}=(\mathbf{P}, \mathbf{Q})$
where $\mathbf{P}, \mathbf{Q} \in \mathbb{H}$. The conjugate, norm, and inverse of the dual quaternion $\mathbf{D}$ is represented as $\mathbf{D}^* = \mathbf{P}^* + \epsilon \mathbf{Q}^*$, $\lVert \mathbf{D} \rVert = \sqrt{\mathbf{D} \mathbf{D}^*} = \sqrt{\mathbf{P} \mathbf{P}^* + \epsilon (\mathbf{P}\mathbf{Q}^* + \mathbf{Q}\mathbf{P}^*)}$, and $\mathbf{D}^{-1} = \mathbf{D}^*/{\lVert \mathbf{D} \rVert}^2$,
% = \mathbf{P}^{-1} (1-\epsilon \mathbf{Q}\mathbf{P}^{-1})
respectively. Another definition for the conjugate of $\mathbf{D}$ is represented as $\mathbf{D}^\dag = \mathbf{P}^* - \epsilon \mathbf{Q}^*$. Addition and multiplication of two dual quaternions $\mathbf{D}_1= \mathbf{P}_1 + \epsilon \mathbf{Q}_1$ and $\mathbf{D}_2= \mathbf{P}_2 + \epsilon \mathbf{Q}_2$ are performed as $\mathbf{D}_1 + \mathbf{D}_2 = (\mathbf{P}_1 + \mathbf{P}_2) + \epsilon (\mathbf{Q}_1 + \mathbf{Q}_2)$ and $\mathbf{D}_1 \mathbf{D}_2 = (\mathbf{P}_1 \mathbf{P}_2) + \epsilon (\mathbf{P}_1 \mathbf{Q}_2 + \mathbf{Q}_1 \mathbf{P}_2) $.
% \orange{As a result, $\mathbf{D}_1 \mathbf{D}_2 \ne \mathbf{D}_2 \mathbf{D}_1$, $(\mathbf{D}_1 \mathbf{D}_2)^* = \mathbf{D}_2^* \mathbf{D}_1^*$, $(\mathbf{D}_1 \mathbf{D}_2)^\dag = \mathbf{D}_2^\dag \mathbf{D}_1^\dag$, and $\mathbf{D}\mathbf{D}^{-1}=\mathbf{D}^{-1}\mathbf{D}=1$.}
The dual quaternion $\mathbf{D}$ is a \textit{unit dual quaternion} if ${\lVert \mathbf{D} \rVert} = 1$, i.e., ${\lVert \mathbf{P} \rVert} = 1$ and $\mathbf{P}\mathbf{Q}^* + \mathbf{Q}\mathbf{P}^* = 0$, and consequently, $\mathbf{D}^{-1} = \mathbf{D}^*$. Unit dual quaternions can be used to represent the group of rigid body displacements, $SE(3) = \mathbb{R}^3 \times SO(3)$, $S E(3)=\left\{(\boldsymbol{R}, \boldsymbol{p}) \mid \boldsymbol{R} \in SO(3), \boldsymbol{p} \in \mathbb{R}^{3}\right\}$. An element $\boldsymbol{T} \in SE(3)$, which is a pose of the rigid body, can also be expressed by a $4 \times 4$ homogeneous transformation matrix as
$\boldsymbol{T} = \left[\begin{smallmatrix}\boldsymbol{R}&\boldsymbol{p}\\\boldsymbol{0}&1\end{smallmatrix}\right]$ where $\boldsymbol{0}$ is a $1 \times 3$ zero vector. A rigid body displacement (or transformation) is represented by a unit dual quaternion $\mathbf{D}_T = \mathbf{Q}_R + \frac{\epsilon}{2} \mathbf{Q}_p \mathbf{Q}_R$ where $\mathbf{Q}_R$ is the unit quaternion corresponding to rotation and  $\mathbf{Q}_p = (0, \boldsymbol{p}) \in \mathbb{H}$ corresponds to the translation. Here, we define $\mathbb{D}$ to represent the set of unit dual quaternions.

%The homogeneous transformation $\boldsymbol{T} \in SE(3)$ (i.e., the rotation $\boldsymbol{R}$ followed by the translation $\boldsymbol{p}$) can be also represented by 

\begin{figure}[!htbp]
    \centering
    \includegraphics[scale=2]{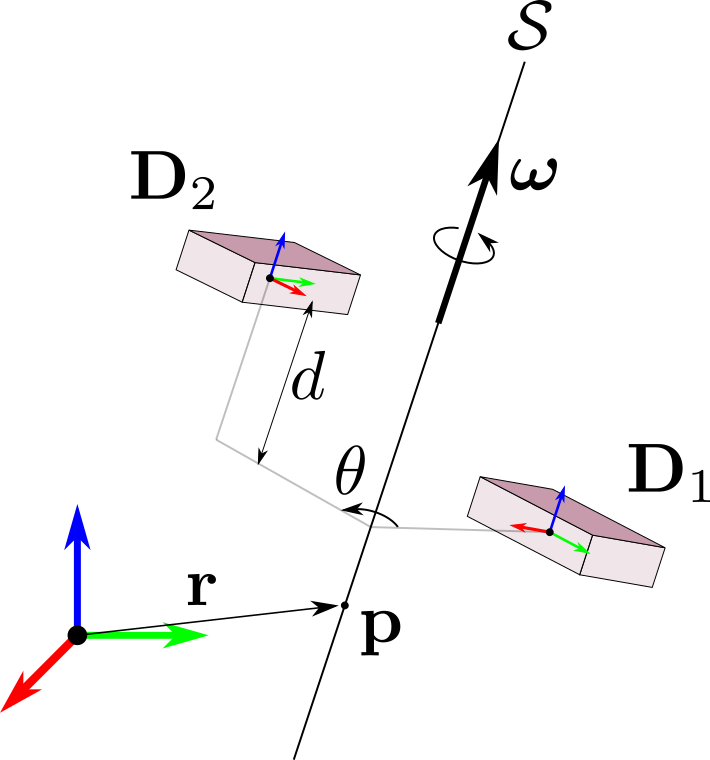}
    \caption{\textbf{\textsc{Screw displacement from pose $\mathbf{D}_1$ to pose $\mathbf{D}_2$}}}
\label{Fig:ScrewDisplacement}
\end{figure}

\noindent
\textbf{Screw Displacement}: Chasles-Mozzi theorem states that the general Euclidean displacement/motion of a rigid body from the origin $\boldsymbol{I}$ to $\boldsymbol{T} = (\boldsymbol{R},\boldsymbol{p}) \in SE(3)$
can be expressed as a rotation $\theta$ about a fixed axis $\mathcal{S}$, called the \textit{screw axis}, and a translation $d$ along that axis (see Fig.~\ref{Fig:ScrewDisplacement}). Plücker coordinates can be used to represent the screw axis by $\boldsymbol{\omega}$ and $\boldsymbol{m}$, where $\boldsymbol{\omega} \in \mathbb{R}^3$ is a unit vector that represents the direction of the screw axis $\mathcal{S}$, $\boldsymbol{m} = \boldsymbol{r} \times \boldsymbol{\omega}$, and $\boldsymbol{r} \in \mathbb{R}^3$ is an arbitrary point on the axis. Thus, the screw parameters are defined as $\boldsymbol{\omega}, \boldsymbol{m}, \theta, h$, where $h = \frac{d}{\theta}$ is the pitch of the screw. In general, $h \geq 0$, with $h = \infty$ for pure translation. When $h$ is finite (infinite), $\theta (d)$ is the magnitude of the screw. A {\em constant screw motion} is a motion where the parameters $\boldsymbol{\omega}, \boldsymbol{m}$, and $h$ stays constant throughout the motion.
% and the pitch of the screw displacement is $\theta/d$.

The screw displacements can be expressed by the dual quaternions as $\mathbf{D}_T = \mathbf{Q}_R + \frac{\epsilon}{2} \mathbf{Q}_p \mathbf{Q}_R = (\cos \frac{\Phi}{2}, L \sin \frac{\Phi}{2})$ where $\Phi = \theta + \epsilon d$ is a dual number and $L = \boldsymbol{\omega} + \epsilon \boldsymbol{m}$ is a
% unit
dual vector.
% with zero scalar part.
A power of the dual quaternion $\mathbf{D}_T$ is then defined as $\mathbf{D}_T^{\tau} = (\cos \frac{\tau \Phi}{2}, L \sin \frac{\tau \Phi}{2})$, $\tau >0$.
% Note that axis $L$ is independent of $\tau$.

% $\mathbf{D}_T = \mathbf{Q}_R + \epsilon \mathbf{Q}_p \mathbf{Q}_R/2 = \cos \frac{\Theta}{2} + L \sin \frac{\Theta}{2}$
% $\mathbf{D}_T^{\tau} = \cos \frac{\tau \Theta}{2} + L \sin \frac{\tau \Theta}{2}$ 

\noindent
\textbf{Screw Linear Interpolation (ScLERP)}: To perform a one degree-of-freedom smooth screw motion (with a constant rotation and translation rate) between two object poses in $SE(3)$, the screw linear interpolation (ScLERP) can be used. The ScLERP provides a \textit{straight line} in $SE(3)$ which is the closest path between two given poses in $SE(3)$. 
%However, note that the origin of the object does not necessarily follow a straight line in
% the real Euclidean space
%$\mathbb{R}^3$.
% If the poses are represented by $4 \times 4$ homogeneous transformation matrices $\boldsymbol{T}_1$ and $\boldsymbol{T}_2$, the path provided by the ScLERP is derived by $\boldsymbol{T}(\tau)=\boldsymbol{T}_{1} \exp \left(\log \left(\boldsymbol{T}_{1}^{-1} \boldsymbol{T}_{2}\right) \tau \right)$ where $ \tau \in[0,1]$ is a scalar path parameter, $\exp(\cdot)$ is matrix exponential, and $\log(\cdot)$ is matrix logarithm. Similarly,
If the poses are represented by unit dual quaternions $\mathbf{D}_{1}$ and $\mathbf{D}_{2}$, the path provided by the ScLERP is derived by $\mathbf{D}(\tau) = \mathbf{D}_1 (\mathbf{D}_1^{-1}\mathbf{D}_2)^{\tau}$ where $ \tau \in[0,1]$ is a scalar path parameter. 
%and $\mathbf{D}_{12} = \mathbf{D}_1^{-1}\mathbf{D}_2 = \mathbf{D}_1^{*}\mathbf{D}_2$ represents the pose $\mathbf{D}_2$ relative to the pose $\mathbf{D}_1$ (or the transformation from $\mathbf{D}_1$ to $\mathbf{D}_2$). 
As $\tau$ increases from 0 to 1, the object moves between two poses along the path
% $\boldsymbol{T}(\tau)$ or
$\mathbf{D}(\tau)$ by the rotation $\tau \theta$ and translation $\tau d$. Let $\mathbf{D}_{12} = \mathbf{D}_1^{-1}\mathbf{D}_2$. To compute $\mathbf{D}_{12}^\tau$, the screw coordinates $\boldsymbol{\omega}, \boldsymbol{m}, \theta, d$ are first extracted from $\mathbf{D}_{12} = \mathbf{P} + \epsilon \mathbf{Q} = (p_0,\boldsymbol{p}_r) + \epsilon (q_0,\boldsymbol{q}_r) = (\cos\frac{\theta}{2}, \boldsymbol{\omega} \sin\frac{\theta}{2}) + \epsilon \mathbf{Q}$ by $\boldsymbol{\omega} = \boldsymbol{p}_r/ \lVert \boldsymbol{p}_r \lVert $, $\theta = 2 \, \mathrm{atan2}(\lVert \boldsymbol{p}_r \lVert, p_0)$, $d = \boldsymbol{p} \cdot \boldsymbol{\omega}$, and $\boldsymbol{m} = \frac{1}{2} (\boldsymbol{p} \times \boldsymbol{\omega} + (\boldsymbol{p}-d \boldsymbol{\omega})\cot \frac{\theta}{2})$ where $\boldsymbol{p}$ is derived from $2\mathbf{Q}\mathbf{P}^* = (0, \boldsymbol{p})$ and $\mathrm{atan2}(\cdot)$ is the two-argument arctangent. Then, $\mathbf{D}_{12}^\tau = (\cos \frac{\tau \Phi}{2}, L \sin \frac{\tau \Phi}{2})$ is directly derived from $\left(\cos \frac{\tau \theta}{2}, \sin \frac{\tau \theta}{2}\boldsymbol{\omega}\right)+\epsilon \left( -\frac{\tau d}{2}\sin \frac{\tau \theta}{2}, \frac{\tau d}{2}\cos \frac{\tau \theta}{2}\boldsymbol{\omega}+\sin \frac{\tau \theta}{2}\boldsymbol{m} \right) $. Note that $h =\infty$ corresponds to pure translation.% and the \textcolor{red}{screw axis is at infinity}.

\noindent
\textbf{Distance Metric in SE(3)}: The heuristic which we follow to determine the distance between two rigid body transformations in space is, to determine the distance in position and orientation separately. Let $\mathbf{D}_1 = \mathbf{Q}_1 + \frac{\epsilon}{2}\mathbf{p}_1 \otimes \mathbf{Q}_1$ and $\mathbf{D}_2 = \mathbf{Q}_2 + \frac{\epsilon}{2}\mathbf{p}_2 \otimes \mathbf{Q}_2$ be two poses in dual quaternion representation.
\begin{itemize}
    \item The distance in position between two poses is given by the Euclidean norm.
    \begin{align*}
    \boldsymbol{d}_{p}(\mathbf{D}_1, \mathbf{D}_2) = || \mathbf{p}_1 - \mathbf{p}_2 ||
    \end{align*}
    \item The distance in orientation between the two poses as defined in \cite{Huynh09} is
    \begin{align*}
        \boldsymbol{d}_{\phi}(\mathbf{D}_1, \mathbf{D}_2) = min\left(||\mathbf{Q}_1 - \mathbf{Q}_2||, ||\mathbf{Q}_1 + \mathbf{Q}_2||\right)
    \end{align*}
\end{itemize}

\noindent
\textbf{Neighbourhood of a pose}: For any pose that is represented in unit dual quaternions as $\mathbf{D}$, we define its $(\varepsilon_p, \varepsilon_\phi)-$neighbourhood as 
\begin{align*}
    \{\mathbf{D'} \in \mathbb{D} | \boldsymbol{d_p}(\mathbf{D}, \mathbf{D'}) \leq \varepsilon_p, \boldsymbol{d_\phi}(\mathbf{D}, \mathbf{D'}) \leq \varepsilon_\phi\}
\end{align*}
The parameters $\varepsilon_p$ and $\varepsilon_\phi$ define the size of the neighbourhood

\section{Problem Statement}
\begin{figure*}
    \centering
    \begin{subfigure}[b]{0.24\textwidth}
        \includegraphics[width=\textwidth]{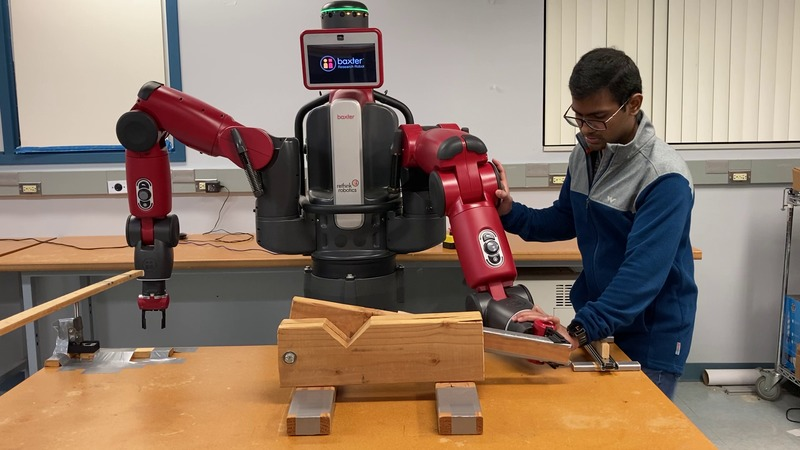}
    \end{subfigure}
    \begin{subfigure}[b]{0.24\textwidth}
        \includegraphics[width=\textwidth]{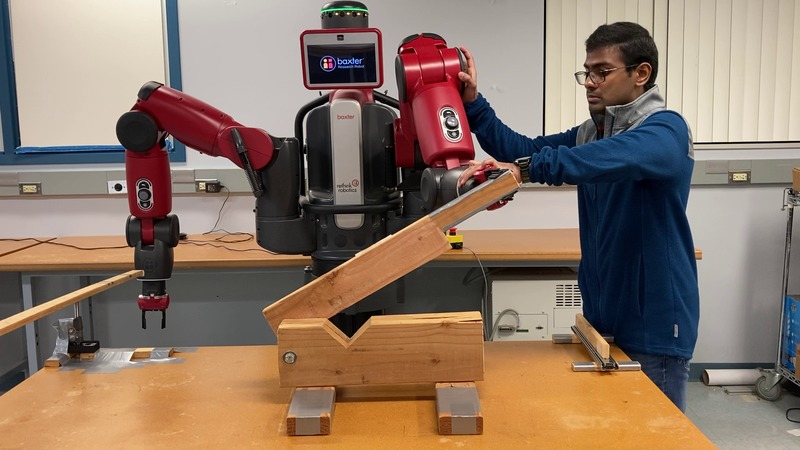}
    \end{subfigure}
    \begin{subfigure}[b]{0.24\textwidth}
        \includegraphics[width=\textwidth]{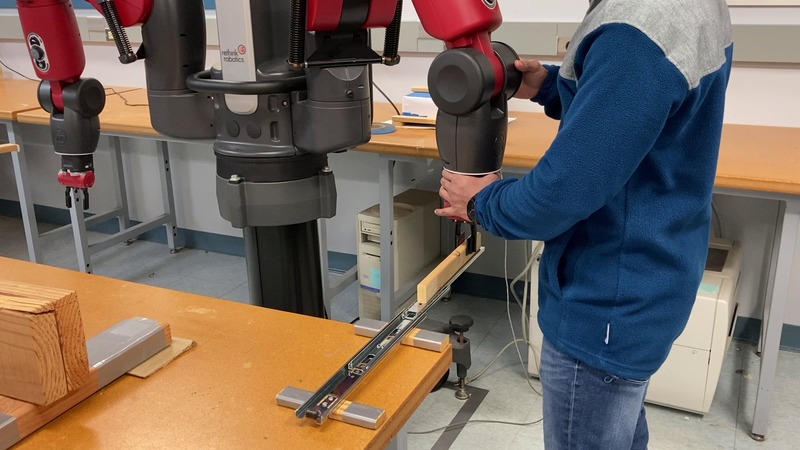}
    \end{subfigure}
    \begin{subfigure}[b]{0.24\textwidth}
        \includegraphics[width=\textwidth]{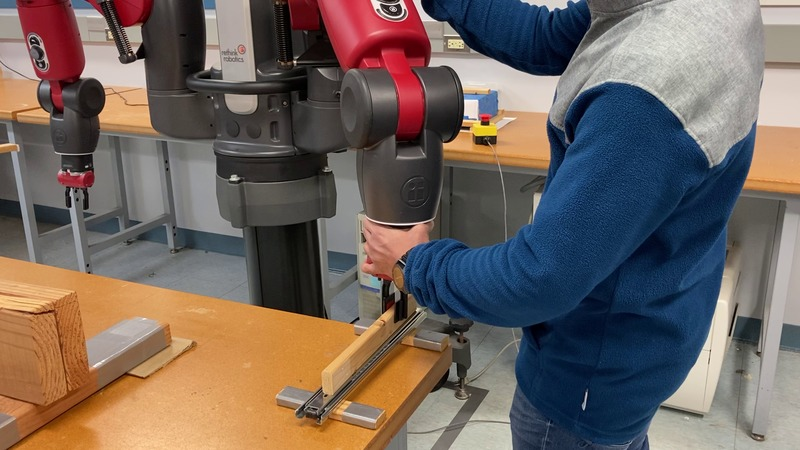}
    \end{subfigure}
    \caption{\textbf{\textsc{Demonstration for Manipulation of Articulated Objects:}} User provided Kinesthetic demonstration of tasks. Demonstration of manipulating an object constrained by a revolute joint (First and Second images from left; Demonstration 3 of Revolute Joint Experiment); Demonstration of manipulation an object constrained by a prismatic joint (Third and Fourth images from left; Demonstration 3 of Prismatic Joint Experiment)}
    \label{fig:articulated_object_demonstration}
\end{figure*}

\begin{figure}[ht!]
    \begin{subfigure}[b]{0.24\textwidth}
        \includegraphics[width=\textwidth]{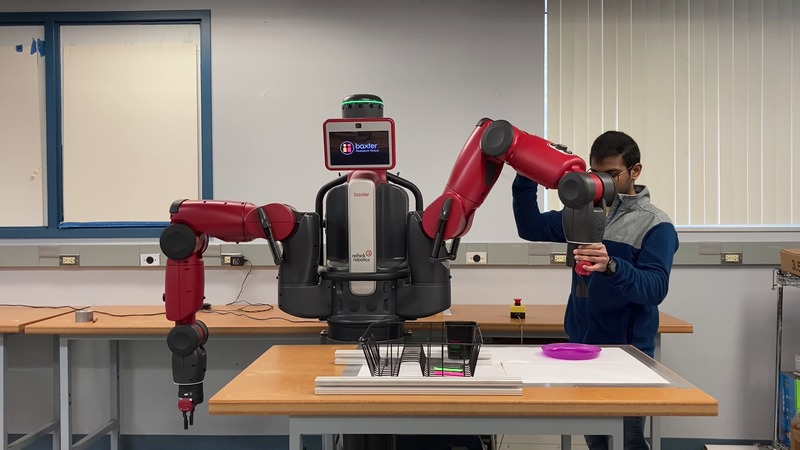}
    \end{subfigure}
    \begin{subfigure}[b]{0.24\textwidth}
        \includegraphics[width=\textwidth]{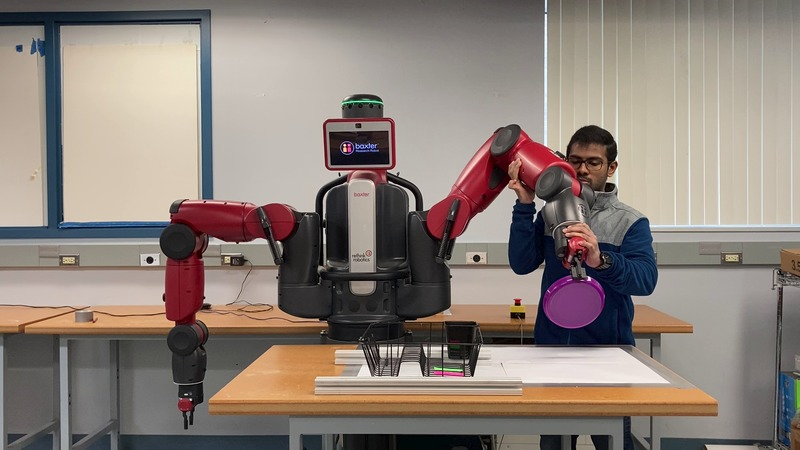}
    \end{subfigure}
    \par\smallskip
    \begin{subfigure}[b]{0.24\textwidth}
        \includegraphics[width=\textwidth]{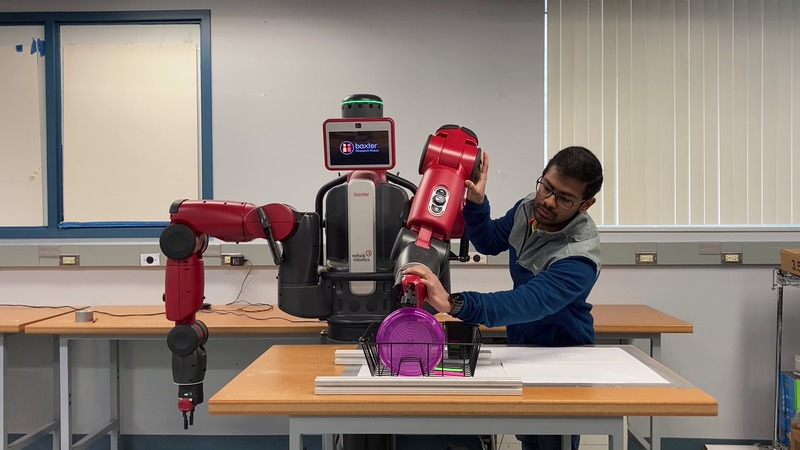}
    \end{subfigure}
    \begin{subfigure}[b]{0.24\textwidth}
        \includegraphics[width=\textwidth]{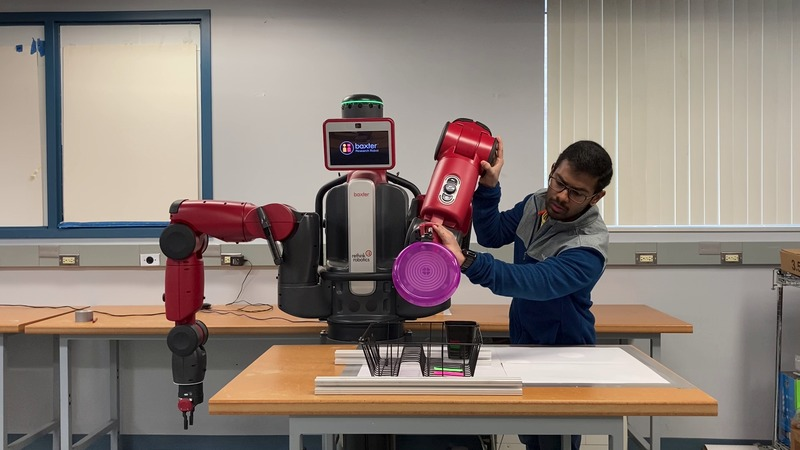}
    \end{subfigure}
    \caption{\textbf{\textsc{Arranging Dishes:}} Demonstration 3\\(Clockwise from top left)}
    \label{fig:arranging_dishes_demonstration}
\end{figure}

Assume that we have a kinesthetic demonstration of a manipulation task where we have recorded the joint angles of the manipulator during the demonstration (See Figure \ref{fig:articulated_object_demonstration}, \ref{fig:scoop_and_pour_demonstration}, \ref{fig:arranging_dishes_demonstration}).
Using the forward kinematics map, we can obtain the sequence of poses the end effector goes through to accomplish the task. Let $\mathcal{D} := \mathbf{D}_1, \mathbf{D}_2, \cdots, \mathbf{D}_n$ be the sequence of task space poses of the end effector and $\mathbf{O}_1, \mathbf{O}_2, ..., \mathbf{O}_v$ be the poses of the task relevant objects written using a unit dual quaternion representation of $SE(3)$. The sequence of end effector poses in $\mathcal{D}$ is the task space representation of the demonstrated path. Any task-relevant constraints that characterize the manipulation task are implicitly present in the demonstration.  Our goal is to find a motion plan for a new initial and final pose $\mathbf{D}_1'$ and $\mathbf{D}_n'$ (as well as the new pose of task-relevant objects $\mathbf{O}_1', \mathbf{O}_2', ..., \mathbf{O}_v'$) that utilizes the single demonstration as a guide to perform the desired task instance and incorporates the task-relevant constraints in the new plan. Note that the use of unit dual quaternion representation is not fundamental here (one can as well use the $4\times4$ transformation matrix representation throughout the whole paper). It is also straightforward and sometimes convenient to go back and forth between the $4 \times 4$ transformation matrix representation and the unit dual quaternion representation of $SE(3)$. 

To accomplish the above, we use a screw geometry-based representation of the underlying demonstrated motion. In particular, we divide the problem above into two sub-problems. 
\begin{enumerate}
    \item {\bf Screw Segmentation Problem}: Given a demonstrated motion in task space and the poses of the task related objects, compute a segmentation of the motion as a sequence of constant screw motions relative to the poses of the task related objects.
    \item {\bf Motion Generation Problem using Segmented Demonstration}: Given a sequence of constant screw motions that is known to accomplish a given task, for a new task instance of the same task (defined by the initial pose and goal pose of the end effector as well as the poses of the task-relevant objects), compute a motion plan that accomplishes the task instance using the sequence of constant screw motions that was determined from the demonstration.
\end{enumerate}
% (b) Given a sequence of constant screw motions that is known to accomplish a given task, for a new task instance of the same task, compute a motion plan that accomplishes the task instance using the given sequence of constant screw motions. \textcolor{red}{Do we also need to know/store the poses of the relevant objects from the demonstration?}

The rationale for using a screw segmentation of the motion is two-fold. First, from Chasles' theorem it can be inferred that any path in $SE(3)$ can be approximated arbitrarily closely as a sequence of constant screw motions. This is analogous to the fact that any curve in $\mathbb{R}^3$ can be approximated arbitrarily closely by a sequence of straight line segments. Second, the screw representation is a coordinate-invariant representation (meaning that it does not depend on the choice of the coordinate frame at the end effector of the robot) and also maps the motion to a single parameter subgroup of $SE(3)$, which potentially allows better generalization properties.

%\noindent
%{\bf Screw Segmentation Problem}:

%\noindent
%{\bf Motion Generation Problem using Segmented Demonstration}:

%$\bf{g}_0^'$ and $\bf{g}_f^'$ that utilizes the demonstration to perform the desired task.
%Our goal is to partition this sequence into a sequence of constant screw motions.  

\section{Screw Segmentation Problem}
\label{sec:screw_parameter_estimation}
In this section, we will discuss the screw segmentation problem and present a solution for the same. Our approach for segmenting the task space motion into constant screws also requires extraction of the screw parameters for each segment. Therefore, for purposes of exposition, we will first discuss the extraction of screws for a motion of the end effector that is generated by a constant screw (e.g., when moving an articulated object like door or drawer whose motion is constrained by a revolute or prismatic joint). Then we will discuss the general case of screw segmentation for a general motion in $SE(3)$. As stated before,  we will use a sequence of recorded end-effector poses in dual quaternion representation, $\mathcal{D} = \{\mathbf{D}_1, \mathbf{D}_2, ..., \mathbf{D}_n\}$.

\noindent
\subsection{\textbf{Sequence Generated by a Single Screw}}
\label{sec:single_screw_parameter_estimation}

The relative transformation between the initial pose $\mathbf{D}_1$ and every other pose on the recorded sequence is given by,
\begin{align}
    \mathbf{D}_{1i} = \mathbf{D}_i \otimes \mathbf{D}_1^*, i = 2,3,...,n
\end{align}
Where, $\mathbf{D}_{1i}$ is of the form $\mathbf{D}_{1i} = \mathbf{Q}_{1i} + \frac{\epsilon}{2}\mathbf{p}_{1i}\otimes\mathbf{Q}_{1i}$.

%Here, $\mathbf{D}_{1i}$ is of the form $\mathbf{D}_{1i} = \mathbf{Q}_{1i} + \frac{\epsilon}{2}\mathbf{p}_{1i}\otimes\mathbf{Q}_{1i}$.
Let us consider the sequence $\mathbf{D}_1, \mathbf{D}_2, ..., \mathbf{D}_n$ to be generated by a single screw with parameters $(\bm{\omega}, \mathbf{m}, h, \theta)$, with the notations as defined before. % $\mathcal{S}$ as the screw axis (defined by the Plücker coordinates $(\bm{\omega}, \mathbf{m})$), $\theta$ as the magnitude and $h$ as the pitch.
If the motion is a general screw motion or pure rotation, then
\begin{gather}
    \mathbf{Q}_{1i} = \bigg(\cos\frac{\theta_{1i}}{2}, \bm{\omega}\sin\frac{\theta_{1i}}{2}\bigg)
    \label{eq:general_screw_orientation_part}\\
    \mathbf{p}_{1i} =
    (\mathbf{I} - e^{\hat{\bm{\omega}}\theta_{1i}})(\bm{\omega}\times\bm{\upsilon})
    + \bm{\omega}\bm{\omega}^T\bm{\upsilon}\theta_{1i}
    \label{eq:general_screw_translation_part}
\end{gather}
where  $\bm{\upsilon} = \mathbf{m} + h\bm{\omega}$.
%\begin{gather}
%    \bm{\upsilon} = \mathbf{m} + h\bm{\omega}
 %   \label{eq:screw_v}
%\end{gather}
%
If the motion is pure translation, then 
%$\mathbf{Q}_{1i} = (1,\bm{0})$ and $\mathbf{p}_{1i} = \bm{\omega}\theta_{1i}$, 
%where $\bm{\omega}=\frac{\mathbf{p}_n-\mathbf{p}_1}{\|\mathbf{p}_n-\mathbf{p}_1\|}$.
\begin{gather}
  \mathbf{Q}_{1i} = (1,\bm{0}) \label{eq:prismatic_orientation_part} \\
   \mathbf{p}_{1i} = \bm{\omega}\theta_{1i} \label{eq:prismatic_translation_part}
\end{gather}
%where $\bm{\omega}=\frac{\mathbf{p}_n-\mathbf{p}_1}{\|\mathbf{p}_n-\mathbf{p}_1\|}$.
where
\begin{gather}
    \bm{\omega}=\frac{\mathbf{p}_n-\mathbf{p}_1}{\|\mathbf{p}_n-\mathbf{p}_1\|}
    \label{eq:prismatic_axis}
\end{gather}

In both cases, the magnitude of the screw is $\theta = \theta_{1n}$, and $\theta_{1i} = \tau_{1i}\theta$, with $ 0 \leq \tau_{1i} \leq 1$ and $\tau_{1i} < \tau_{1j}$ for $i < j$. Thus, the entire motion is a one-parameter motion, where the parameter is denoted by $\tau_{1i}$ and each end-effector pose corresponds to a different  $\tau_{1i}$.

\noindent
\textbf{Noiseless Observations: }We will first discuss the situation where we assume that there is no noise in the observation. This is for exposition purposes only, so that the basic idea can be clearly explained. In the subsequent discussion we will look at the noisy case.
In the noiseless case, we can take any index $i$, i.e., any intermediate pose on the recorded path to determine the type of screw motion by computing $\mathbf{Q}_{1i}$ with Equation \eqref{eq:general_screw_orientation_part} and \eqref{eq:prismatic_orientation_part}. Based on the type of screw motion, we use either Equation \eqref{eq:general_screw_orientation_part} or \eqref{eq:prismatic_translation_part} to solve for the screw axis $\bm{\omega}$. Note that the value of $\theta_{1i}$ obtained from \eqref{eq:general_screw_orientation_part} or \eqref{eq:prismatic_translation_part} will depend on the chosen index $i$ (and it is not part of the screw). The value of $\bm{\omega}$ will be independent of the index chosen and it should be the same for every index $i \geq 2$.

For pure translation, the pitch, $h = \infty$ and $\bm{m} = \bm{0}$ by definition.
For general motion, we can first obtain $\bm{\omega}$ from Equation~\eqref{eq:general_screw_orientation_part} and $\bm{\upsilon}$ from Equation \eqref{eq:general_screw_translation_part} using the following expression 
\begin{equation}
\label{eq:screw_v}
   \bm{\upsilon}  =
    \left[(\mathbf{I} - e^{\hat{\bm{\omega}}\theta_{1i}})\hat{\bm{\omega}}
    + \theta_{1i}\bm{\omega}\bm{\omega}^T\right]^{-1}\mathbf{p}_{1i}
\end{equation}
where the $3 \times 3$ matrix $\left[(\mathbf{I} - e^{\hat{\bm{\omega}}\theta_{1i}})\hat{\bm{\omega}}
    + \theta_{1i}\bm{\omega}\bm{\omega}^T\right]$ is always invertible. Then we get get the pitch, $h$, of the screw using 
\begin{equation}
  \label{eq:screw_pitch}
  h = \bm{\omega}^T\bm{\upsilon}
\end{equation}
and we get $\mathbf{m}$ by using $\mathbf{m} = \bm{\upsilon} - h\bm{\omega}$.
If $h = 0$, then the motion is pure rotation. 
%To obtain $\mathbf{m}$, we first obtain  Using $\bm{\upsilon}$, we get $\mathbf{m} = \bm{\upsilon} - h\bm{\omega}$.

\noindent
\textbf{Noisy Observations: }In the presence of noise in observations (which occurs in practical scenarios), the screw parameters extracted from each intermediate pose $i$ may be different. 
As the screw parameter determination is different for pure translation and general screw/rotational motion, we compute the screw parameters for the motion between the initial pose $\mathbf{D}_1$ and final pose $\mathbf{D}_n$ by first verifying which of the following hypothesis about the screw motion category are true
\begin{itemize}
    \item Pure translation ($h = \infty$)
    \item General screw/pure rotation ($h \neq \infty$)
\end{itemize}

%We use the computed screw parameters, to then determine which of the two hypothesis is true. 
%We first try to determine if the motion is pure translation. If so we compute

Ideally, if the sequence $\{\mathbf{D}_1, \mathbf{D}_2, ..., \mathbf{D}_{n-1}, \mathbf{D}_n\}$ has been generated by a constant screw, then $\forall ~i \leq n$, there is a $\theta_{1k}$ such that, 
$\mathbf{D}_k = \mathbf{D}_{1k} \otimes \mathbf{D}_n$,
where $\mathbf{D}_{1k} = \mathbf{Q}_{1k} + \frac{\epsilon}{2}\mathbf{p}_{1k} \otimes \mathbf{Q}_{1k}$,
and has screw parameters $(\bm\omega_{1n}, \mathbf{m}_{1n}, h_{1n}, \theta_{1k})$.
But, due to the presence of noise, $\mathbf{D}_k$ does not lie on the ideal screw motion from $\mathbf{D}_1$ to $\mathbf{D}_n$. However, using line search on the parameter $\tau$, $0 \leq \tau \leq 1$ we can determine a $\theta_{1k} = \tau\theta_{in}$ which gives us 
$\mathbf{D}_k' = \mathbf{D}_{1k} \otimes \mathbf{D}_1$
such that $\mathbf{D}_k'$ lies on the screw motion from $\mathbf{D}_1$ to $\mathbf{D}_n$ and within a neighbourhood $(\varepsilon_p, \varepsilon_\phi)$ of the pose $\mathbf{D}_k$. If no such $\theta_{1k}$ exists then $\mathbf{D}_k$ does not lie on the screw motion from $\mathbf{D}_1$ to $\mathbf{D}_n$. We will be using this to verify if the given sequence $\mathcal{D}$ can be generated from a constant screw and also to check which of the two hypothesis is true.

\begin{algorithm}[t!]
\RestyleAlgo{ruled}
\caption{Check if motion is pure translation}\label{alg:check_if_prismatic}
\SetKwFunction{checkIfPrismatic}{checkIfPrismatic}
\SetKw{Continue}{continue}
\SetKwProg{Fn}{def}{\string:}{}

\Fn{\checkIfPrismatic{$\mathcal{D}, \varepsilon_p, \varepsilon_{\phi}$}}{
    Determine $\mathbf{p}_{1n}$ from $\mathbf{D}_{1n} = \mathbf{D}_n \otimes \mathbf{D}_1^*$\\
    %$\bm{\omega}=\frac{\mathbf{p}_n-\mathbf{p}_1}{\|\mathbf{p}_n-\mathbf{p}_1\|}, \theta=\|\mathbf{p}_n-\mathbf{p}_1\|$\\
    Compute $\theta, \bm{\omega}$ from $\mathbf{p}_{1n}$ using \eqref{eq:prismatic_translation_part}, \eqref{eq:prismatic_axis}\\
    \For{$i = 2$ \KwTo $n-1$}
    {
        \eIf{\textnormal{$\tau_{1i}$ exists such that $\mathbf{D}_k'$ lies in the neighborhood $(\varepsilon_p, \varepsilon_{\phi})$ of $\mathbf{D}_k$}}
        {\Continue}
        {\Return false}
    }
    
    \Return true
}
\end{algorithm}

\begin{algorithm}[t!]
\RestyleAlgo{ruled}
\caption{Check if motion is pure rotation or general screw}\label{alg:check_if_rotation_general_screw}
\SetKwFunction{checkIfGeneralScrew}{checkIfGeneralScrew}
\SetKw{Continue}{continue}
\SetKwProg{Fn}{def}{\string:}{}

\Fn{\checkIfGeneralScrew{$\mathcal{D}, \varepsilon_p, \varepsilon_{\phi}$}}{
    %$\mathbf{Q}_{1n} = \bigg(\cos\frac{\theta_{1n}}{2}, \bm{\omega}\sin\frac{\theta_{1n}}{2}\bigg)$\\
    %$\mathbf{p}_{1n} =
    %(\mathbf{I} - e^{\hat{\bm{\omega}}\theta_{1n}})(\bm{\omega}\times\bm{\upsilon})
    %+ \bm{\omega}\bm{\omega}^T\bm{\upsilon}\theta_{1n}$\\
    Determine $\mathbf{Q}_{1n}$ and $\mathbf{p}_{1n}$ from $\mathbf{D}_{1n} = \mathbf{D}_n \otimes \mathbf{D}_1^*$\\ %using \eqref{eq:general_screw_orientation_part} and \eqref{eq:general_screw_translation_part}\\
    Get $\theta, \bm{\omega}$ from $\mathbf{Q}_{1n}$, and $h, \mathbf{m}$ from $\mathbf{p}_{1n}$ using  \eqref{eq:screw_pitch} and \eqref{eq:screw_v}\\
    \For{$i = 2$ \KwTo $n-1$}
    {
        \eIf{\textnormal{$\tau_{1i}$ exists such that $\mathbf{D}_k'$ lies in the neighborhood $(\varepsilon_p, \varepsilon_{\phi})$ of $\mathbf{D}_k$}}
        {\Continue}
        {\Return false}
    }
    
    \Return true
}
\end{algorithm}

\begin{algorithm}[t!]
\RestyleAlgo{ruled}
\caption{Determination of Screw Parameters}\label{alg:screw_param_determination}
\DontPrintSemicolon
\SetKwFunction{getScrewParameters}{getScrewParameters}
\SetKwFunction{checkIfPrismatic}{checkIfPrismatic}
\SetKwFunction{checkIfGeneralScrew}{checkIfGeneralScrew}
\SetKwProg{Fn}{def}{\string:}{}

\Fn{\getScrewParameters{$\mathcal{D}, \varepsilon_p, \varepsilon_{\phi}$}}{
    $\mathbf{D}_{1n} = \mathbf{D}_n \otimes \mathbf{D}_1^*$\\
    \uIf{\checkIfPrismatic{$\mathcal{D}, \varepsilon_p, \varepsilon_{\phi}$}\textnormal{\textbf{is} true}}
    {
        Compute $\bm{\omega}, \theta$ from $\mathbf{D}_{1n}$ using \eqref{eq:prismatic_translation_part}, $h = \infty, \mathbf{m} = \bm{0}$\\
        \Return $\bm{\omega}, \mathbf{m}, \theta, h$
    }
    \uElseIf{\checkIfGeneralScrew{$\mathcal{D}, \varepsilon_p, \varepsilon_{\phi}$}\textnormal{\textbf{is} true}}
    {
        Compute $\bm{\omega}, \theta, h, \mathbf{m}$ from $\mathbf{D}_{1n}$ using \eqref{eq:general_screw_orientation_part}, \eqref{eq:screw_pitch}, \eqref{eq:general_screw_translation_part}, \eqref{eq:screw_v}\\
        \Return $\bm{\omega}, \mathbf{m}, \theta, h$
    }
    \Else
    {
        \Return $\mathcal{D}$ is not generated by a single screw
    }
}

\end{algorithm}

Algorithm \ref{alg:check_if_prismatic} gives the procedure to check if the given motion $\mathcal{D}$ belongs to the category of pure translation ($h = \infty$). Given the motion $\mathcal{D}$ and the parameters $\varepsilon_p, \varepsilon_{\phi}$ to define the size of the neighborhood, the algorithm returns true if the motion $\mathcal{D}$ is pure translation and false otherwise. The screw parameters are computed from the initial pose $\mathbf{D}_1$ and final pose $\mathbf{D}_n$ and then used to determine if each of the intermediate poses $\mathbf{D}_i, 2 \leq i \leq n-1$ lie within a neighborhood $(\varepsilon_p, \varepsilon_{\phi})$ of the ideal screw motion from $\mathbf{D}_1$ to $\mathbf{D}_n$ inside the loop from lines 4 to 8.

Algorithm \ref{alg:check_if_rotation_general_screw} is similar to Algorithm \ref{alg:check_if_prismatic} but gives the procedure to check if the given motion $\mathcal{D}$ belongs to the category of general screw or pure rotational motion ($h \neq \infty$).

Algorithm \ref{alg:screw_param_determination} gives the procedure to compute the constant screw parameters of the given motion $\mathcal{D}$. It takes as input the sequence of poses $\mathcal{D}$ and the parameters $\varepsilon_p, \varepsilon_{\phi}$ to define the size of the neighborhood, and returns the constant screw parameters if the motion $\mathcal{D}$ was generated by a constant screw motion. It first checks if the given motion is pure translation and if it is true then computes the screw parameters as mentioned in line 4. If the motion is not pure translation, but a general screw or pure rotation, then it computes the screw parameters as mentioned in line 7. If the previous two conditions are not satisfied, then the given motion $\mathcal{D}$ is not generated by a constant screw motion.

It is important to note that due to the presence of noise, motion that results from prismatic motion might be fit to a general screw with $h \neq \infty$ depending on choice of the parameters $\varepsilon_p$ and $\varepsilon_\phi$. Also, in section \ref{sec:screw_segmentation}, we will be showing how we can extend this check to segment any given sequence of $SE(3)$ poses into a sequence of constant screws.

\subsection{\textbf{Segmentation of a Sequence of Screws}}
\label{sec:screw_segmentation}
If the sequence $\mathcal{D}$ was generated by a sequence of constant screw motions, then we can represent the motion as a sequence of screw displacements 
%$\{(\mathcal{S}_1,h_1,\theta_1), (\mathcal{S}_2,h_2,\theta_2), ..., (\mathcal{S}_x,h_x,\theta_x)\}$
$\bm\delta_1, \bm\delta_2, ..., \bm\delta_u$
applied to the initial pose $\mathbf{D}_1$ where $1 \leq u < n$, or as a sequence of rigid body poses $\{ \textbf{E}_1, \textbf{E}_2, ...,\textbf{E}_u\}$ where $\textbf{E}_1 = \bm\delta_1 \otimes \mathbf{D}_1$ and $\textbf{E}_i = \bm\delta_i \otimes \mathbf{E}_{i-1}$ for $i = 2,...,u$. Here, $\mathbf{E}_i$ represents the end pose of each constant screw segment in sequence and $u$ represents the number of constant screws present in the motion $\mathcal{D}$. If $u = 1$, then this reduces to the case of single screw motion discussed previously in Section \ref{sec:single_screw_parameter_estimation}. While in some cases the representation based on screw parameters is useful, in other cases the representation based on $SE(3)$ poses is useful and we will be going back and forth between these representations in this paper.

As stated before, according to Chasles Theorem, any rigid body displacement between two given poses can be expressed as a constant screw motion. So given two poses $\mathbf{D}_1$ and $\mathbf{D}_2$, the rigid body displacement can be realised by the screw motion $\bm{\delta} = \mathbf{D}_2 \otimes \mathbf{D}_1^*$ with screw parameters $(\bm\omega, \mathbf{m}, h, \theta)$ and any pose $\mathbf{D}_\tau$ on the screw motion between $\mathbf{D}_1$ and $\mathbf{D}_2$ can be computed using the parameters $(\bm\omega, \mathbf{m}, h, \tau\theta), ~0 \leq \tau \leq 1$.

Ideally, if the sequence $\{\mathbf{D}_i, \mathbf{D}_{i+1}, ..., \mathbf{D}_{j-1}, \mathbf{D}_j\}$ has been generated by a constant screw, then $\forall ~i < k \leq j ~\exists ~\theta_k$ such that, 
$\mathbf{D}_k = \mathbf{D}_{ik} \otimes \mathbf{D}_i$,
where $\mathbf{D}_{ik} = \mathbf{Q}_{ik} + \frac{\epsilon}{2}\mathbf{p}_{ik} \otimes \mathbf{Q}_{ik}$,
with screw parameters $(\bm\omega_{ij}, \mathbf{m}_{ij}, h_{ij}, \theta_{ik})$.
But, due to the presence of noise, $\mathbf{D}_k$ does not lie on the ideal screw motion from $\mathbf{D}_i$ to $\mathbf{D}_j$. However, as discussed before, we can determine a $\theta_{ik}$ which gives us 
%$\mathbf{D}_k' = (\cos{\frac{\theta_{ik}+\epsilon h_{ij} \theta_{ik}}{2}},
%(\bm\omega_{ij} + \epsilon\mathbf{m}_{ij})\sin{\frac{\theta_{ik}+\epsilon h_{ij} \theta_{ik}}{2}}) \otimes \mathbf{D}_1$
%such that 
$\mathbf{D}_k'$ that lies within a neighbourhood $(\varepsilon_p, \varepsilon_\phi)$ of the pose $\mathbf{D}_k$.
If no such $\theta_k$ exists then $\mathbf{D}_k$ does not lie on the screw motion from $\mathbf{D}_i$ to $\mathbf{D}_j$ and this means that the sequence $\{\mathbf{D}_i, \mathbf{D}_{i+1}, ..., \mathbf{D}_{j-1}, \mathbf{D}_j\}$ is not generated by a constant screw.

This is the same check we discussed previously in Section \ref{sec:single_screw_parameter_estimation} and using this procedure as a test to determine whether the given sequence of $SE(3)$ poses is generated by a constant screw or not, we determine the indices $i,j$ for all the $u$ constant screw segments present in the sequence $\mathcal{D}$.

\begin{comment}

\begin{algorithm}[t!]
\RestyleAlgo{ruled}
\caption{Check if given motion is generated by a constant screw}\label{alg:check_if_screw}
\SetKwFunction{checkIfScrew}{checkIfScrew}
\SetKw{Continue}{continue}
\SetKwProg{Fn}{def}{\string:}{}

\Fn{\checkIfScrew{$\mathcal{D}, \varepsilon_p, \varepsilon_{\phi}$}}{
    \uIf{\checkIfPrismatic{$\mathcal{D}, \varepsilon_p, \varepsilon_{\phi}$}\textnormal{\textbf{is} true}}
    {
        \Return true
    }
    \uElseIf{\checkIfGeneralScrew{$\mathcal{D}, \varepsilon_p, \varepsilon_{\phi}$}\textnormal{\textbf{is} true}}
    {
        \Return true
    }
    \Else
    {
        \Return false
    }
}
\end{algorithm}

Algorithm \ref{alg:check_if_screw} determines if the given sequence of $SE(3)$ poses $\mathcal{D}$ was generated by a constant screw motion. It checks if all the poses in $\mathcal{D}$ are within a certain neighbourhood $(\varepsilon_p, \varepsilon_{\phi})$ of the ideal screw motion from $\mathbf{D}_1$ to $\mathbf{D}_n$.

\end{comment}

\begin{algorithm}[t!]
\RestyleAlgo{ruled}
\caption{Segment given motion into a sequence of constant screws}\label{alg:get_screw_segs}
\SetKwFunction{getScrewSegments}{getScrewSegments}
\SetKwFunction{checkIfScrew}{checkIfScrew}
\SetKwFunction{getScrewParameters}{getScrewParameters}
\SetKw{Continue}{continue}
\SetKw{Break}{break}
\SetKwProg{Fn}{def}{\string:}{}

\Fn{\getScrewSegments{$\mathcal{D}, \varepsilon_p, \varepsilon_{\phi}$}}{
    Set $u = 1, i = 1, \mathbf{E}_u = \mathbf{D}_n$\\
    \While{$i <= n$}
    {
        \For{$j = i+1$ \KwTo $n$}
        {
            $\mathcal{D}' = \{\mathbf{D}_i, ..., \mathbf{D}_j\}$\\
            \eIf{
            %\checkIfScrew{$\mathcal{D}', \varepsilon_p, %\varepsilon_{\phi}$}\textnormal{\textbf{is} true}
            %\uIf{
            \getScrewParameters{$\mathcal{D}', \varepsilon_p, \varepsilon_{\phi}$}
            \textnormal{\textbf{returns}} $\bm{\omega}, \mathbf{m}, \theta, h$
            }
            {
                \Continue
            }
            %\uElseIf{$\mathcal{D}'$ \textnormal{is not generated by a single screw}}
            {
                $\mathbf{E}_u$ = $\mathbf{D}_{j-1}$\\
                $u = u + 1$\\
                $i = j$\\
                \Break
            }
        }
    }
    
    \Return $\{\mathbf{E}_1, \mathbf{E}_2, ..., \mathbf{E}_u\}$
}
\end{algorithm}

%Algorithm \ref{alg:get_screw_segs} uses the screw motion check implemented in Algorithm  \ref{alg:check_if_screw} to determine 
Algorithm \ref{alg:get_screw_segs} uses Algorithm \ref{alg:screw_param_determination} to determine the start pose $\mathbf{D}_i$ and end pose $\mathbf{D}_j$ of each screw segment in the given $SE(3)$ pose sequence such that all the intermediate poses $\mathbf{D}_{i+1}, ..., \mathbf{D}_{j-1}$ can be fit within a given tolerance $(\varepsilon_p, \varepsilon_{\phi})$ to the ideal screw from $\mathbf{D}_i$ to $\mathbf{D}_j$.

The output of Algorithm \ref{alg:get_screw_segs} is the segmentation of the motion of the end effector in the task space as a sequence of constant screws. This sequence of constant screws describe the task constraints in a coordinate invariant manner, i.e., it does not depend on the choice of the end effector coordinate frame. By extracting the task constraints which are enforced either due to the presence of joints or due to the inherent nature of the task, we are able to generalize a single demonstration to different task instances using the planning method described in Section \ref{sec:motion_generation}.

\section{Motion Generation from Segmented Demonstration}
\label{sec:motion_generation}
In this section, we will describe our method of motion generation based on the description of a demonstrated motion as a sequence of screw segments (along with the poses of the objects, if any). As stated before, we consider manipulation tasks belonging to two different classes. First, we consider manipulation of articulated objects, which is usually accomplished by a single constant screw motion. 
%Examples of such tasks include opening/closing a door (or manipulating any object with a revolute joint) and opening/closing a drawer (or manipulating any object with a prismatic joint). 
Second, we consider complex manipulation tasks that can be executed by a sequence of constant screw motions. 
%Examples of such task includes, transferring granular objects from one container to another and loading a dishwasher. Thus, formally, we define a gross manipulation task as a sequence of constant screw motions.

%Given a kinesthetic demonstration of a task, we segment the motion of the end effector in the task space into a sequence of constant screws. This sequence of constant screws describes the task constraints in a coordinate invariant manner. By extracting the task constraints which are enforced either due to the presence of joints or due to the inherent nature of the task, we are able to generalize a single demonstration to different task instances.

\begin{figure*}[t!]
    \centering
    \includegraphics[width=0.95\textwidth]{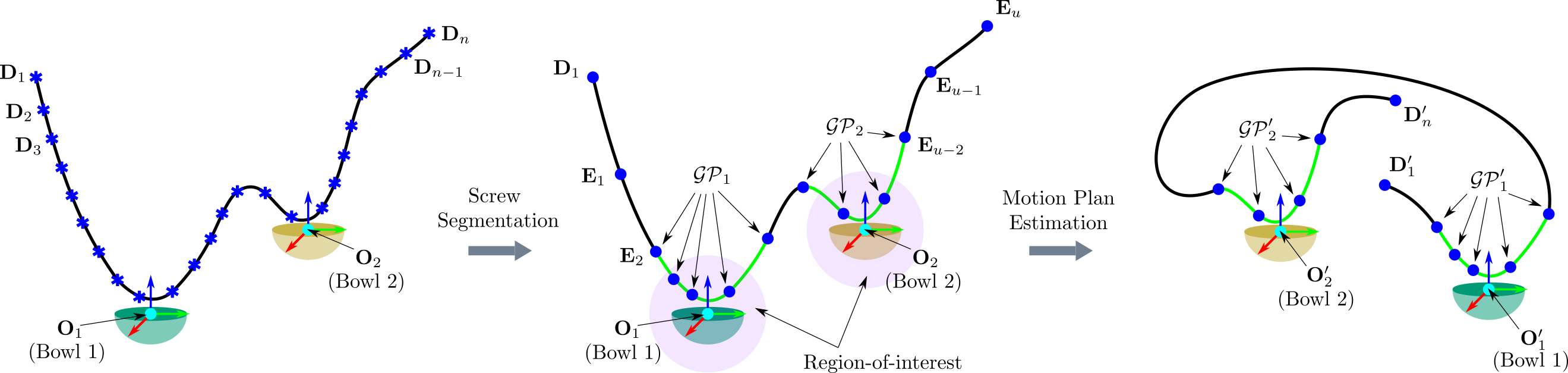}
    \caption{\textbf{\textsc{Schematic Sketch for Motion Estimation of Complex Task}:} Overview of motion estimation for a scoop and pour task. The task involves scooping contents from Bowl 1 with a spoon and pouring it into Bowl 2. The $SE(3)$ poses are represented as points to reduce clutter; \textbf{Left }- The provided demonstration $\mathcal{D}$ consisting of a sequence of $SE(3)$ poses along with pose of the objects $\mathbf{O}_1$ (Bowl 1) and $\mathbf{O}_2$ (Bowl 2); \textbf{Center }- Segmenting the provided demonstration $\mathcal{D}$ into a sequence of constant screws $\{\mathbf{D}_1, \mathbf{E}_1, ..., \mathbf{E}_u\}$ and determining the \textbf{Key Segments} (Coloured Green) associated with each object. The task-relevant constraints $\mathcal{GP}_1$ (Bowl 1), $\mathcal{GP}_2$ (Bowl 2) are identified from the Key Segments; \textbf{Right }- Computing the motion plan for new initial pose $\mathbf{D}_1'$, final pose $\mathbf{D}_2'$ and object poses $\mathbf{O}_1'$ (Bowl 1) and $\mathbf{O}_2'$ (Bowl 2) by determining the \textbf{Guiding Poses} $\mathcal{GP}'$ utilizing the task-relevant constraints (Coloured Green) obtained from the previous step}
    \label{fig:Complex_Manipulation_Schematic}
\end{figure*}
\begin{comment}
\begin{figure*}[t!]
    \centering
    \begin{subfigure}[b]{0.45\textwidth}
        \includegraphics[width=\textwidth]{figures/Plots/ArrangingDishes/ArrangingDishesScrewSegments_D3.png}
    \end{subfigure}
    \begin{subfigure}[b]{0.45\textwidth}
        \includegraphics[width=\textwidth]{figures/Plots/ArrangingDishes/ArrangingDishesKeySegments_D3.png}
    \end{subfigure}
    \par
    \includegraphics[width=0.35\textwidth]{figures/Plots/ArrangingDishes/Legend.png}
    \caption{\textbf{\textsc{Recorded Demonstration from Arranging Dishes Task Experiment:}} Recorded end-effector trajectory as a sequence of SE(3) poses from an actual demonstration. Majority of the recorded $SE(3)$ poses are hidden to reduce clutter; \textbf{Left} - Segmentation of the provided demonstration $\mathcal{D}$ into a sequence of constant screws. The constant screw segments are differentiated using different colours; \textbf{Right} - The \textbf{Key Segments} are coloured green and the $\mathcal{GP}_i$ poses are represented using Magenta coloured markers}
    \label{fig:determination_of_task_related_constraints}
\end{figure*}
\end{comment}
\subsection{\textbf{Manipulation of Articulated Objects}}
\noindent
\textbf{Problem Statement: }\textit{Given a single demonstration $\mathcal{D} = \{\mathbf{D}_1, \mathbf{D}_2, ..., \mathbf{D}_n\}$ for manipulating articulated objects, determine the motion plan required to manipulate the articulated object for a new initial pose $\mathbf{D}_1'$ of the end-effector and magnitude of motion $\theta'$}

Articulated objects are constrained in their motion based on the type of joint (revolute/prismatic) present. Given the screw parameters of the joints, we can compute the motion of the end-effector required to manipulate the objects. We determine these parameters $(\bm\omega, \mathbf{m}, h, \theta)$ from $\bm{\delta} = \mathbf{D}_n \otimes \mathbf{D}_1^*$.

%The recorded demonstration, due to the constrained nature of the task, consists of a single screw segment whose parameters define the joint constraints. We can segment the demonstration and check if the number of screw segments is equal to one to ensure that the provided demonstration involved manipulating articulated objects. Choosing very small values for the screw segmentation parameters $\varepsilon_p$ and $\varepsilon_\phi$ sometimes results in multiple segments due to the presence of noise in the recordings. Also, if there is any error in the demonstration provided by the user the number of segments will be greater than one. For a good demonstration which is not overwhelmed with noise, we can determine the screw parameters $(\bm\omega, \mathbf{m}, h, \theta)$ from $\bm{\delta} = \mathbf{D}_n \otimes \mathbf{D}_1^*$.
%Segmenting the recorded demonstration usually results in a single screw segment whose parameters $(\bm\omega, \mathbf{m}, h, \theta)$ define the joint constraints.
Here $\theta$ is just the magnitude of the screw motion and can be changed depending on the direction or the magnitude by which we want to move the object. For example, we can vary the angle by which we open/close the door about its hinge using $\theta$. Once we have estimated the screw parameters, given a new initial pose $\mathbf{D}_1'$ of the end-effector and magnitude of motion $\theta'$, we can use the screw parameters to compute the screw displacement
$\bm\delta' = (\cos{\frac{\theta'+\epsilon h \theta'}{2}},
(\bm\omega + \epsilon\mathbf{m})\sin{\frac{\theta'+\epsilon h \theta'}{2}})$
and the new final configuration $\mathbf{D}_n' = \bm\delta' \otimes \mathbf{D}_1'$. Once we have determined the goal pose $\mathbf{D}_n'$, we can compute the motion plan from $\mathbf{D}_1'$ to $\mathbf{D}_n'$ which is required to manipulate the articulated object using the ScLERP motion planner \cite{sclerp_motion_planner} so that the motion satisfies the screw constraint. Note that our approach does not need {\em a priori} information about the articulation model to estimate the joint constraints. Further, the fact that ScLERP will generate poses of the end-effector that satisfies the constraint imposed by the joint without explicitly considering it is proven in~\cite{sclerp_motion_planner}. 
%The coordinate-invariant of the screw representation allows us to obtain this result simply 

\subsection{\textbf{Complex Manipulation Tasks}}
\label{sec:motion_generation_complex_manipulation_task}
\noindent
\textbf{Problem Statement: }\textit{Given a single demonstration of a task $\mathcal{D} = \{\mathbf{D}_1, \mathbf{D}_2, ..., \mathbf{D}_n\}$ and the poses $\{\mathbf{O}_{1}, \mathbf{O}_{2}, ..., \mathbf{O}_{v}\}$ of the $v$ task-relevant objects during the demonstration, compute the motion plan required to perform the task for new poses for the initial pose, $\mathbf{D}_1'$ and final pose, $\mathbf{D}_n'$ and the task-relevant objects $\{\mathbf{O}_{1}', \mathbf{O}_{2}', ..., \mathbf{O}_{v}'\}$}

For complex manipulation tasks we segment the provided demonstration into a sequence of constant screws which can then be used to determine the task constraints and generate the motion plan for a new task instance. %Given the demonstration $\bf{D}_0, \bf{D}_1, \cdots, \bf{D}_f$, we segment it into a sequence of constant screws that can be represented as a sequence of rigid body poses $\{\textbf{D}_1, \textbf{E}_1, \textbf{E}_2, ...,\textbf{E}_x\}$ where each pose defines the start of a new screw segment and the end of the previous screw segment.

\noindent
\textbf{Determination of Task Relevant Constraints: }
%\textit{We first determine the task constraints as a sequence of guiding poses relative to the pose of the objects}
The task-relevant constraints that are implicitly present in the provided demonstration are approximated as a sequence of constant screws. Based on a heuristic that the task-relevant constraints occur in a region-of-interest surrounding the task-relevant objects, we define the task-relevant constraints as \textbf{\textit{the sequence of screw segments obtained from the provided demonstration such that they lie inside the region-of-interest surrounding the task-related objects and expressed with respect to the local frame of reference of the associated task-related object}}. Each task-relevant object has its own region-of-interest and a sequence of screws associated with it. Let us refer to the screw segments that lie inside a region-of-interest as \textbf{\textit{key segments}}. Here, we define the region-of-interest of a task-related object to be a spherical volume or a cuboid centered at the object frame with dimensions that is set based on the type of the object. However it can also be chosen to be any other shape that captures the motion that is associated with the task-relevant objects.

We segment the provided demonstration $\mathcal{D}$ into a sequence of $u$ constant screws $\{\textbf{E}_1, \textbf{E}_2, ...,\textbf{E}_u\}, ~u \leq (n-1)$ (We will be using SE(3) poses to represent screw segments). Then, we construct the regions-of-interest for all the $v$ task-relevant objects at $\{\mathbf{O}_{1}, \mathbf{O}_{2}, ..., \mathbf{O}_{v}\}$ and determine the key segments associated with each object.
%(Figure \ref{fig:determination_of_task_related_constraints}).

The key-segments associated with each task related object $i$ are determined as $\{\mathbf{G}_{i,1}, \mathbf{G}_{i,2}, ..., \mathbf{G}_{i,m_i}\}$, where $m_i$ is the number of key-segments associated with the object.

The task-relevant constraints for object $i$ expressed in its local frame of reference are then computed as,
\begin{align}
\label{eq:task_constraints_from_demonstration}
%\mathcal{GP}_i = \{\mathbf{G}_{i,1} \otimes \mathbf{O}_i^*, \mathbf{G}_{i,2} \otimes \mathbf{O}_i^*, ..., \mathbf{G}_{i,m_i} \otimes \mathbf{O}_i^*\}
\mathcal{GP}_i = \{\mathbf{O}_i^* \otimes \mathbf{G}_{i,1}, \mathbf{O}_i^* \otimes \mathbf{G}_{i,2}, ..., \mathbf{O}_i^* \otimes \mathbf{G}_{i,m_i}\}
\end{align}

\noindent
\textbf{Generation of Motion Plan for a new task instance: }
Given the new poses of the task related objects, $\{\mathbf{O}_{1}', \mathbf{O}_{2}', ..., \mathbf{O}_{v}'\}$, we can re-compute the task-relevant constraints with respect to the new pose of the object.

For the new pose $\mathbf{O}_i'$ of each object $i$, the task-relevant constraints can be computed from \eqref{eq:task_constraints_from_demonstration} as
\begin{align}
\mathcal{GP}_i' = \{\mathbf{O}_i' \otimes \mathbf{O}_i^* \otimes \mathbf{G}_{i,1}, ..., \mathbf{O}_i' \otimes \mathbf{O}_i^* \otimes \mathbf{G}_{i,m_i}\}  
\end{align}

%We use the pose of the task related objects along with its associated guiding poses determined using the demonstration to estimate the motion plan required to perform a new instance of the task. Given the new poses of the task related objects, using the relative change in pose of the objects with respect to their pose during the demonstration, we recompute the guiding poses associated with each object (Figure \ref{fig:determination_of_motion_plan_for_another_task_instance}).

\begin{figure}[h!]
    \centering
    \includegraphics[width=0.45\textwidth]{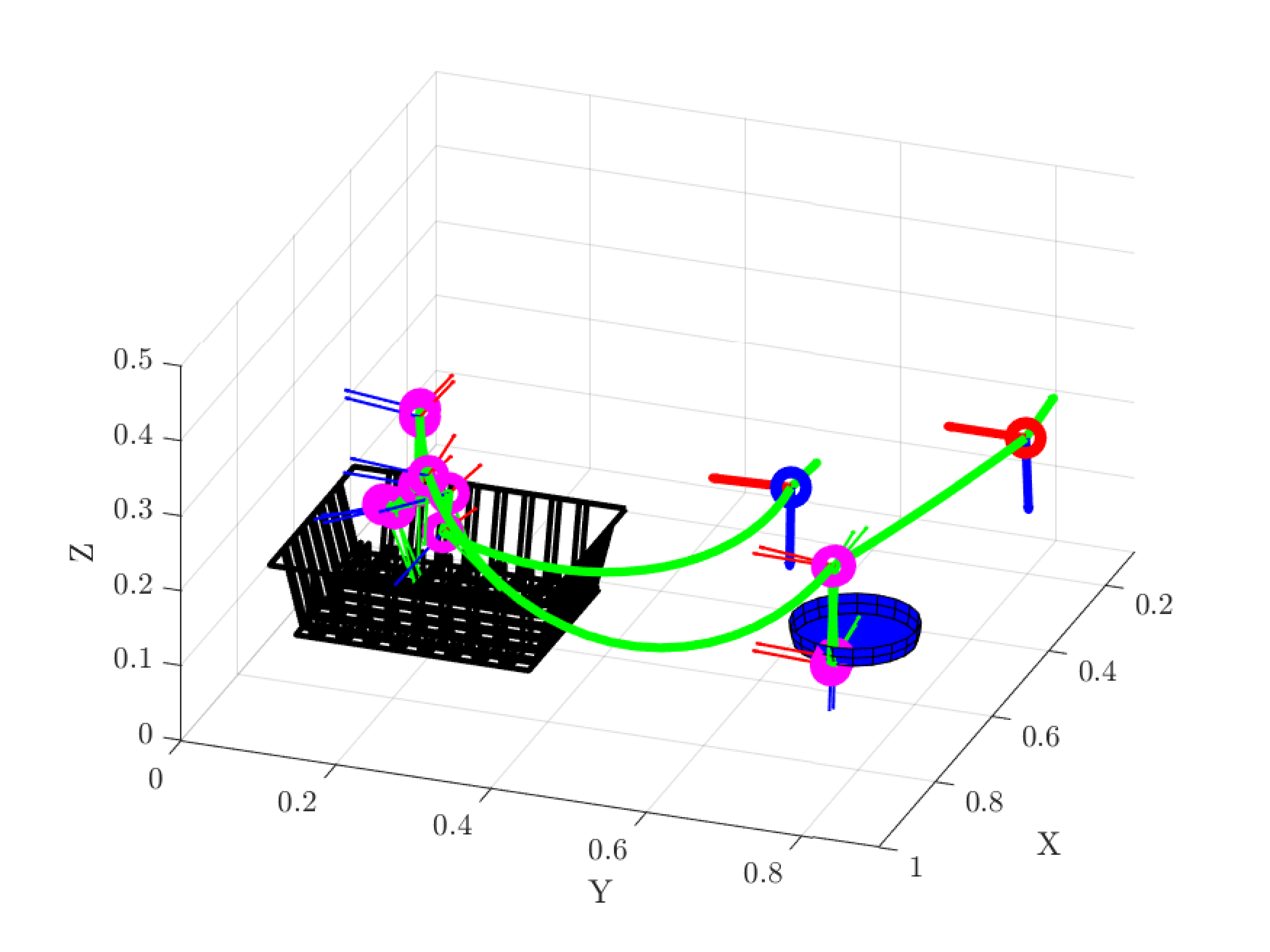}
    \par
    \includegraphics[width=0.35\textwidth]{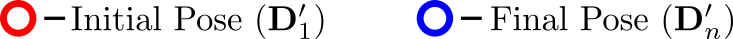}
    \caption{\textbf{\textsc{Motion Plan for a new task instance of the Arranging Dishes Task:}} The motion plan is computed by determining the \textbf{Guiding Poses}, $\mathcal{GP'}$ (Magenta coloured markers) as detailed in Section \ref{sec:motion_generation_complex_manipulation_task} for the new initial pose, final pose and object poses. The final motion (Green coloured trajectory) is obtained by interpolating using ScLERP between consecutive guiding poses}
    \label{fig:determination_of_motion_plan_for_another_task_instance}
\end{figure}

\begin{comment}
\begin{figure}[h!]
    \begin{subfigure}[b]{0.24\textwidth}
        \includegraphics[width=\textwidth]{figures/Plots/ArrangingDishes/DishesInterpolatedTrajectory.jpg}
    \end{subfigure}
    \begin{subfigure}[b]{0.24\textwidth}
        \includegraphics[width=\textwidth]{figures/Plots/ArrangingDishes/DishedNewMotionPlanTrajectory.jpg}
    \end{subfigure}
    \caption{Determination of motion plan using guiding poses for same instance (left) and for another instance (right) of the Arranging Dishes task}
    \label{fig:determination_of_motion_plan_for_another_task_instance}
\end{figure}
\end{comment}

Once the task-relevant constraints for the new task instance have been computed, we construct the sequence of constant screws that the end-effector should follow to successfully execute the task. We define this sequence as the \textbf{\textit{Guiding Poses}}. The guiding poses can be constructed as,
\begin{align}
    \mathcal{GP}' = \{\mathbf{D}_1', \mathcal{GP}_1', \mathcal{GP}_2', ..., \mathcal{GP}_v', \mathbf{D}_n'\}
\end{align}

The guiding poses $\mathcal{GP}'$ define the motion plan for the given new instance of the task. Using the ScLERP Motion Planner \cite{sclerp_motion_planner} to move to each of the poses in $\mathcal{GP}'$ ensures that the task-relevant constraints are satisfied while the motion is being executed.

\section{Experimental Results}
In this section we provide experimental results for four manipulation tasks that can be classified into two categories: (a) manipulation of articulated objects, and (b) complex manipulation tasks. For category (a), we consider two tasks: (1) manipulation of an object constrained by a revolute joint (2) manipulation of an object constrained by a prismatic joint, and for category (b), we consider two tasks: (3) scooping and pouring (4) arranging dishes in a rack. We conducted a total of $118$ experimental trials over the four manipulation tasks.
%obtained by using our approach to generate a motion plan for new instances of a demonstrated task. 

The experiments are performed using the Baxter robot from Rethink Robotics~\cite{baxter_hardware_specs}. A demonstration is provided using a kinesthetic interface. During a demonstration, the joint angles are recorded, and by using the forward kinematics map of the robot, we obtain the sequence of poses in $SE(3)$, which is segmented to compute the sequence of constant screw motions that capture the task constraints. More information on the experiments have been included in the supplementary video\footnote{\url{https://youtu.be/Ss9MQ-6nIDE}}.

%We consider four manipulation tasks that can be classified into two categories: (a) manipulation of articulated objects, and (b) complex manipulation tasks. For category (a), we consider two tasks: (1) manipulation of an object constrained by a revolute joint (2) manipulation of an object constrained by a prismatic joint, and for category (b), we consider two tasks: (3) scooping and pouring (4) arranging dishes in a rack.

%\textcolor{red}{Define a metric for success/failure of each task}

\subsection{\textbf{Manipulation of Articulated Objects}}
For each task, we test our algorithm on $3$ different demonstrations.
%through the entire range of motion of the object. 
For each demonstration, the gripper held the object in a different place with a different pose. For each demonstration, we perform $13$ experimental trials with the extracted screw, where the gripper pose varied across the different trials and were different from that in the demonstration. Thus, we conducted a total of $78$ experimental trials ($39$ for each task). The purpose of varying the gripper poses is to show that, practically, the coordinate invariance implies that irrespective of how we hold the object, we can use the extracted screw from just a single demonstration to manipulate the object. 
%and then use this demonstration to determine the screw constraints. After we have determined the screw constraints, we change the grasping pose of the gripper and then make the robot perform the same task through the entire range of motion of the object based on its new grasping pose. 

%The recorded demonstration, due to the constrained nature of the task, consists of a single screw segment whose parameters define the joint constraints. If the data is very noisy or if there was any user introduced error during the demonstration, then segmenting the demonstration will result in multiple screw segments even for the constrained manipulation task. Essentially, screw segmentation can be used to verify if the provided motion is constrained or not by looking at the number of screw segments.
%If we set the values of $\varepsilon_p$ and $\varepsilon_\phi$ too low during screw segmentation, our algorithm becomes too sensitive to the noise and starts to over-fit. At the worst case scenario, our algorithm will compute the number of screw segments, $u= n-1$, where $n$ is the number of $SE(3)$ poses recorded during the demonstration.
%Setting the value of $\varepsilon_p$ and $\varepsilon_\phi$ too high during screw segmentation, results in under-fitting as the screw segment is fit based on only the initial ($\mathbf{D}_1$) and final ($\mathbf{D}_n)$ poses and any information present in the intermediate poses (\{$\mathbf{D}_2, ..., \mathbf{D}_{n-1}\})$ are dropped out.

Recall that we had two parameters in our segmentation algorithm, namely, $\varepsilon_p$ and $\varepsilon_\phi$.
Across all our experiments, we set the values of the parameters as $\varepsilon_p = 1 \text{ cm}$ and $\varepsilon_{\phi} = 0.1$.
%resulted in a single screw segment when segmenting demonstrations of manipulating constrained objects. Reducing $\varepsilon_p$ and $\varepsilon_\phi$ below the above values increased the number of screw segments to more than one.

\subsubsection{\textbf{Object constrained by a Revolute Joint}}

The object used for this task is a wooden block constrained to rotate about a revolute joint. Figure~\ref{fig:articulated_object_demonstration} shows the object and snapshots of one demonstration. 
%The wooden block is constrained to only rotate about its revolute axis. 
For this experiment, although the demonstrations were for opening the wooden block, the robot was made to perform the task of both opening and closing it. 
The pose of the gripper was changed between each trial by moving the gripper along the length of the handle and also changing the orientation of the gripper. 
This task was split into two test cases, namely, (a) opening by an angle of $45\degree$ from its closed position, and (b) closing by an angle of $45\degree$ from its open position.

\begin{figure}[h!]
    \begin{subfigure}[b]{0.24\textwidth}
        \includegraphics[width=\textwidth]{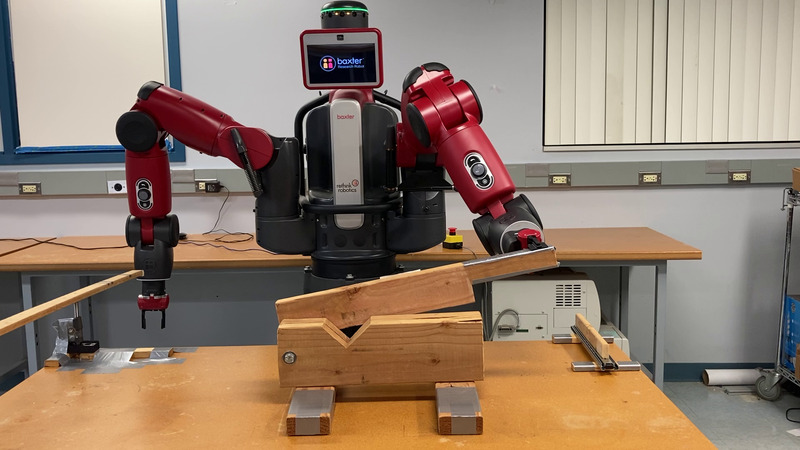}
        \caption{Trial 2}
    \end{subfigure}
    \begin{subfigure}[b]{0.24\textwidth}
        \includegraphics[width=\textwidth]{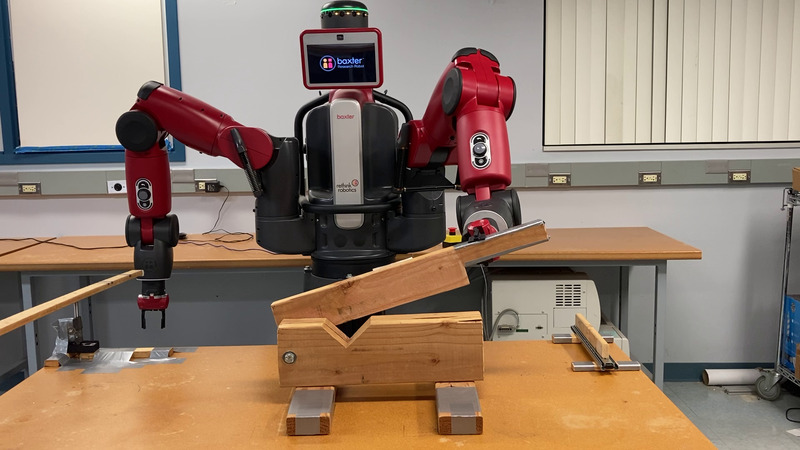}
        \caption{Trial 4}
    \end{subfigure}
    \caption{\textbf{\textsc{Grasping Pose Changes - Revolute Joint Experiment:}} Changes in position and orientation of the end-effector along the handle of the wooden block between different trials using Demonstration 3}
    \label{fig:revolute_joint_grasping_pose_changes}
\end{figure}

%\begin{itemize}
%    \item Opening by an angle of 45\degree~from its closed position
%    \item Closing by an angle of 45\degree~from its open position
%\end{itemize}
%This experiment was performed for three iterations. During each iteration of the experiment, a demonstration of opening the wooden block from the closed position was provided and using this single demonstration 13 trials were performed. 
For each of the $3$ demonstrations, the task of opening was performed $8$ times and the task of closing was performed $5$ times. Figure~\ref{fig:revolute_joint_grasping_pose_changes} shows the change of the gripper poses for different trials.
%A total of three demonstrations were provided, each consisting of only opening the wooden block from the closed position. For each demonstration, the task of opening was performed 8 times with random grasping poses and the task of closing was performed 5 times with random grasping poses.
For all the trials, the robot was able to successfully open/close the wooden block by the given magnitude of $45\degree$ about the revolute joint. 
%Even though the position and the orientation at which the robot grasps the handle of the wooden block was changed between the provided demonstration and each individual trial (See Figure \ref{fig:revolute_joint_grasping_pose_changes}, the robot was able to successfully open and close the wooden block.
%
%\textcolor{blue}{The table is not important here and can be removed. The important thing to discuss here is that during each of the trials the pose of the end effector was different from the one shown in the trials. However all trials were successful, which shows the invariance of the approach to the end effector pose when holding the object. In other words, we can easily generalize across different poses of holding the handle. The pictures showing different holding poses of holding are useful, but it should be explicitly mentioned in the figure caption. The other important aspect here is generalization to closing, which is already mentioned. What is the sensitivity of your result to the parameters chosen? Did you study that? Whether we studied it or not, we need to make a comment about the choice of the parameters. }
Even though all the provided demonstrations were of opening the block, by using the extracted screw parameters with only a change in the sign of the magnitude of the screw, we were also able to perform the task of closing the block.
\begin{comment}
\begin{figure}[h!]
    \begin{subfigure}[b]{0.24\textwidth}
        \includegraphics[width=\textwidth]{figures/RevoluteJoint/RevoluteJointDemo3Trial3Start.jpg}
        \caption{Start pose (Closed Position)}
    \end{subfigure}
    \begin{subfigure}[b]{0.24\textwidth}
        \includegraphics[width=\textwidth]{figures/RevoluteJoint/RevoluteJointDemo3Trial3End.jpg}
        \caption{End pose (Open Position)}
    \end{subfigure}
    \caption{\textbf{\textsc{Revolute Joint:}} Trial 3 using Demonstration 3}
    \label{fig:revolute_joint_demo_3_trial_3}
\end{figure}
\end{comment}

\begin{figure}[h!]
    \begin{subfigure}[b]{0.24\textwidth}
        \includegraphics[width=\textwidth]{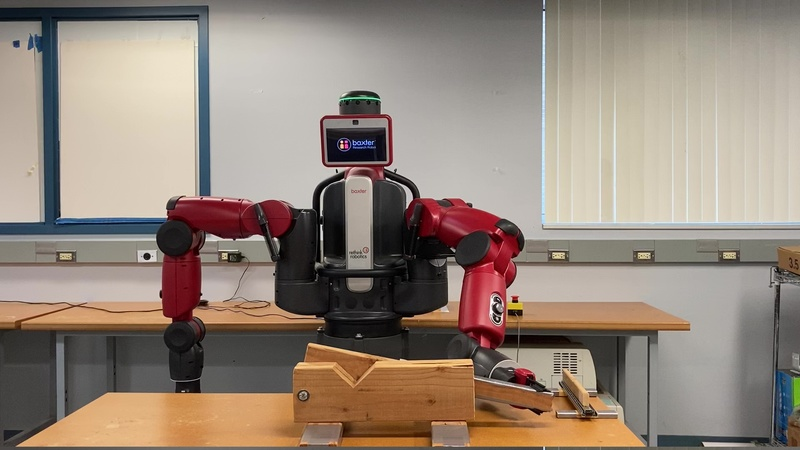}
        \caption{Start pose (Closed Position)}
    \end{subfigure}
    \begin{subfigure}[b]{0.24\textwidth}
        \includegraphics[width=\textwidth]{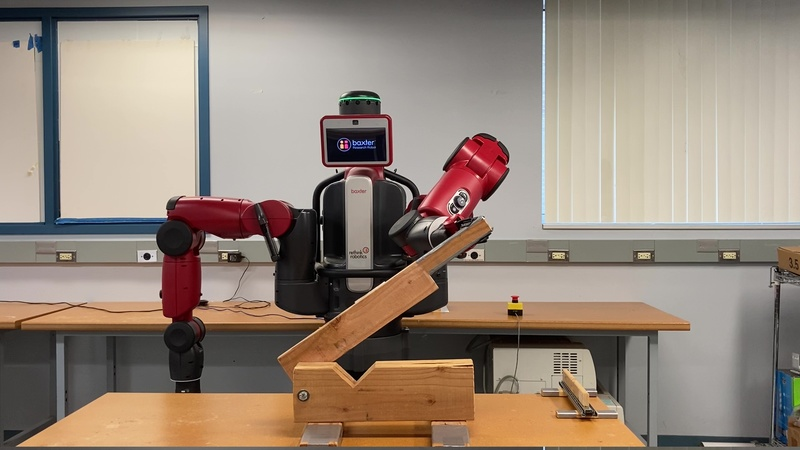}
        \caption{End pose (Open Position)}
    \end{subfigure}
    \caption{\textbf{\textsc{Revolute Joint:}} Trial 7 using Demonstration 3}
    \label{fig:revolute_joint_demo_3_trial_7}
\end{figure}

\begin{table}[h!]
\centering
\begin{tabular}{||c|c|c|c||} 
    \hline
    \hline
    \textbf{Demo} & 
    $\bm{\omega}$ & 
    $\bm{\theta}$ (deg)& 
    $h$\\
    \hline
    \hline
    1 &
    $\begin{bmatrix} 0.9992 & 0.0264 & -0.0293 \end{bmatrix}^T$
    & 41.01 & 0.0202\\
    \hline
    2 &
    $\begin{bmatrix} 0.9997 & 0.0188 & -0.0152 \end{bmatrix}^T$
    & 39.97 & 0.0117\\
    \hline
    3 &
    $\begin{bmatrix} 0.9881 & 0.1439 & 0.0551\end{bmatrix}^T$
    & 38.36 & 0.0053\\
    \hline
    \hline
\end{tabular}
\caption{Revolute Joint - Screw Parameters computed from recorded demonstrations}
\label{revolute_joint_screw_parameters}
\end{table}

Table~\ref{revolute_joint_screw_parameters} shows some relevant screw parameters for the three demonstrations of moving the wooden block constrained by a revolute joint. The second column shows the screw axis, which in this case is the same as the axis of the revolute joint. Note that the computation has been done using the poses of the gripper for which both the position and the orientation changes. However, the pitch (shown in the last column) is almost $0$, which shows that the motion is generated by a pure rotation. The slight deviation from $0$ in the estimated pitch is due to the presence of noise in the joint encoders and clearance in the revolute joint. Furthermore, the estimated axis of the revolute joint is almost identical (within the noise range).  

\subsubsection{\textbf{Object constrained by a Prismatic Joint}}
\begin{figure}[h!]
    \begin{subfigure}[b]{0.24\textwidth}
        \includegraphics[width=\textwidth]{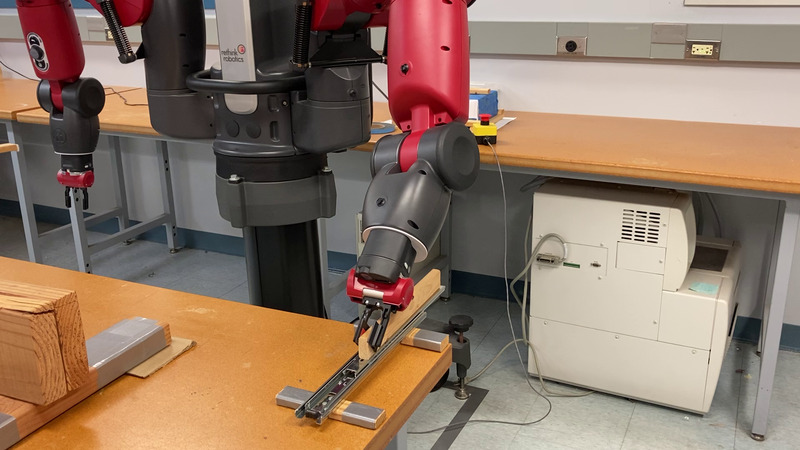}
        \caption{Trial 3}
    \end{subfigure}
    \begin{subfigure}[b]{0.24\textwidth}
        \includegraphics[width=\textwidth]{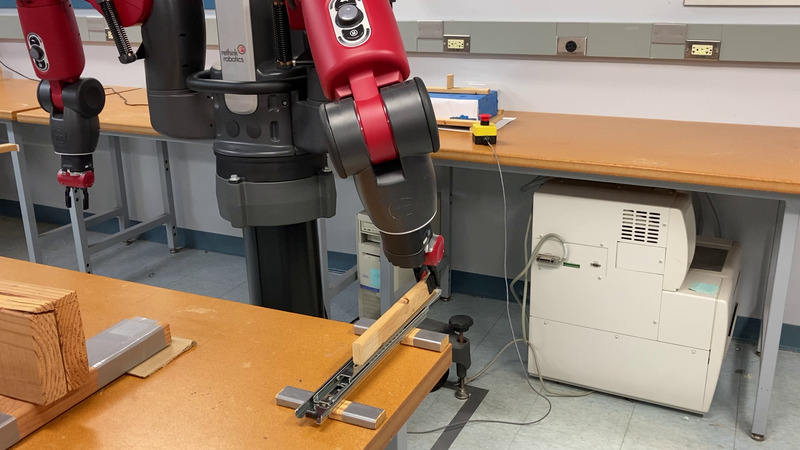}
        \caption{Trial 4}
    \end{subfigure}
    \caption{\textbf{\textsc{Grasping Pose Changes - Prismatic Joint Experiment:}} Changes in position and orientation of the end-effector between different trials using Demonstration 3}
    \label{fig:prismatic_joint_grasping_pose_changes}
\end{figure}

The object used for this task is a wooden block constrained to translate along a prismatic joint axis (see Fig. \ref{fig:articulated_object_demonstration}). %The wooden block is constrained to only translate about its prismatic axis. For this experiment the screw parameters were determined form the provided kinesthetic demonstration and the robot was made to perform the tasks of both opening and closing the block. The grasping pose of the end effector was changed for each demonstration and also in-between each trial conducted using a demonstration. 
This task was also categorized into two cases, namely, (a) opening from the closed position by a distance of $30$ cm, and (b) closing from the open position by a distance of $30$ cm.
%\begin{itemize}
%    \item Opening from the closed position by a distance of 30 cm
%    \item Closing from the open position by a distance of 30 cm
%\end{itemize}
%This experiment was performed for three iterations. During each iteration of the experiment, a single demonstration was provided and using this demonstration, 13 trials were conducted.
Among the $3$ demonstrations, the first and the second was that of opening the block from the closed position while the third demonstration was that of closing the block from the open position. For each demonstration, opening was performed $8$ times and closing the block was performed $5$ times with a random grasping pose for each trial (See Figure \ref{fig:prismatic_joint_grasping_pose_changes}).

\begin{figure}[ht!]
    \begin{subfigure}[b]{0.24\textwidth}
        \includegraphics[width=\textwidth]{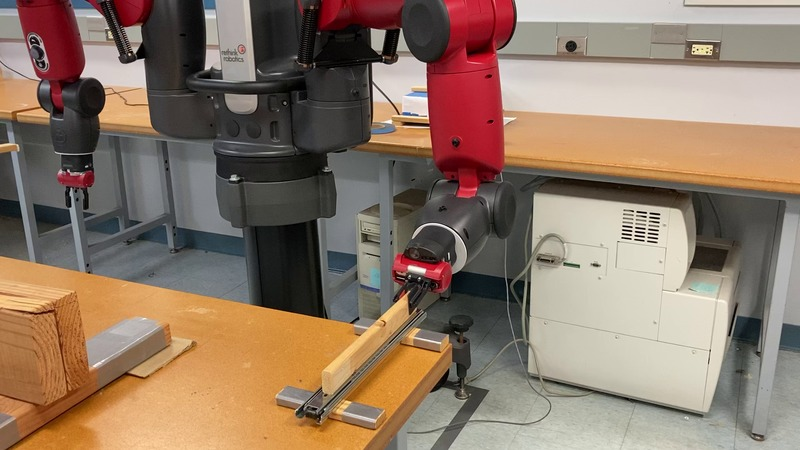}
        \caption{Start pose (Closed Position)}
    \end{subfigure}
    \begin{subfigure}[b]{0.24\textwidth}
        \includegraphics[width=\textwidth]{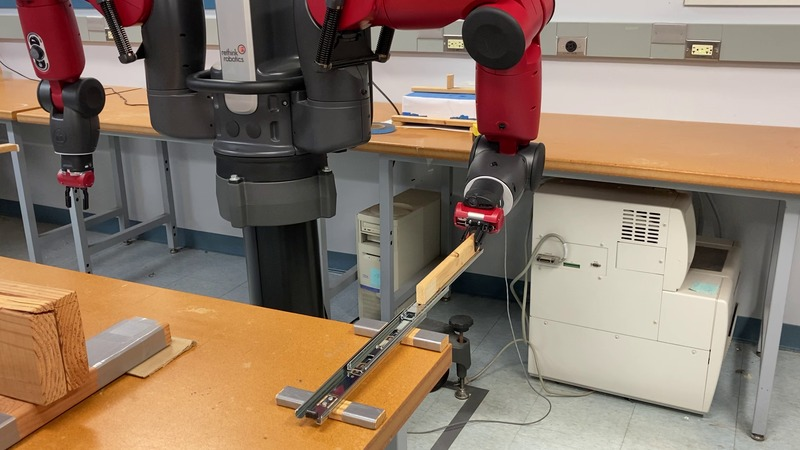}
        \caption{End pose (Open Position)}
    \end{subfigure}
    \caption{\textbf{\textsc{Prismatic Joint:}} Trial 2 using Demonstration 3}
    \label{fig:prismatic_joint_demo_3_trial_2}
\end{figure}
\begin{comment}
\begin{figure}[ht!]
    \begin{subfigure}[b]{0.24\textwidth}
        \includegraphics[width=\textwidth]{figures/PrismaticJoint/PrismaticJointDemo3Trial6Start.jpg}
        \caption{Start pose}
    \end{subfigure}
    \begin{subfigure}[b]{0.24\textwidth}
        \includegraphics[width=\textwidth]{figures/PrismaticJoint/PrismaticJointDemo3Trial6End.jpg}
        \caption{End pose}
    \end{subfigure}
    \caption{\textbf{\textsc{Prismatic Joint:}} Trial 7 using Demonstration 3}
    \label{fig:prismatic_joint_demo_3_trial_7}
\end{figure}
\end{comment}
All of the experimental trials were successful and as before changing the sign of the screw magnitude allowed us to accomplish both opening and closing tasks even though the demonstration may be of the opposite type. %Similar to the previous experiment, irrespective of the demonstration being either of opening the wooden block or closing the wooden block, since we are computing the screw parameters, we can perform the closing task using the demonstration provided for opening and vice-versa. 
Here again, the importance of using the screw geometry of motion is highlighted by the fact that all the $39$ trials were performed successfully irrespective of the pose at which the end effector grasped the object (See Figure \ref{fig:prismatic_joint_grasping_pose_changes}).

%It is important to note that due to noise present in the recorded demonstration, data recorded from a purely transnational motion could also be fit to a general screw ($h \neq \infty$) and the motion can be performed successfully performed with the general screw parameters. However, it is better to classify the motion as pure translation using the procedure described in Algorithm \ref{alg:screw_param_determination} and perform the motion using the pure translation screw parameters.
%\textcolor{blue}{Same comment as before. The table does not provide any information. The generalization with respect to the holding poses is the key bit of information here. The screw parameters here would be interesting, since computationally the pitch cannot be infinity. Did we use the knowledge that the joint is prismatic? }

\begin{table}[h!]
\centering
\begin{tabular}{||c|c|c|c||} 
    \hline
    \hline
    \textbf{Demo} & 
    $\bm{\omega}$ & 
    $\bm{\theta}$ (m)\\
    \hline
    \hline
    1 &
    $\begin{bmatrix} -0.9997 & -0.0169 & 0.0199 \end{bmatrix}^T$
    & 0.3270\\
    \hline
    2 &
    $\begin{bmatrix} -0.9989 & -0.0147 & 0.0452 \end{bmatrix}^T$
    & 0.3207\\
    \hline
    3 &
    $\begin{bmatrix} 0.9995 & 0.0257 & -0.0184 \end{bmatrix}^T$
    & 0.3209\\
    \hline
    \hline
\end{tabular}
\caption{Prismatic Joint - Screw Parameters computed from recorded demonstrations}
\label{prismatic_joint_screw_parameters}
\end{table}

Table~\ref{prismatic_joint_screw_parameters} shows some of the screw parameters for the prismatic joint obtained from the three demonstrations. Here, the estimated screw axis, $\bm{\omega}$ is same as the axis of the prismatic joint. The estimated joint axis only vary slightly due to the presence of noise in the joint encoders and the clearance in the joint. Note that the estimated joint axis for the third demonstration  is in the opposite direction compared to the first two demonstrations. This is because, in the third demonstration, we were closing (or pushing) the block, whereas we were opening (or pulling) the block in the first two demonstrations.

\subsection{\textbf{Complex Manipulation Tasks}}
For complex manipulation tasks, along with the kinesthetic demonstration, we also record the poses of the task related objects. We segment the end-effector path into a sequence of constant screws and extract the key segments. During task execution, we vary the pose of the task-related objects. Given the new poses of the objects, we compute the guiding poses and use those to determine the new motion plan. The values of the screw segmentation parameters for complex manipulation tasks was set as $\varepsilon_p = 1 \text{ cm}$ and $\varepsilon_{\phi} = 0.15$.

\subsubsection{\textbf{Scooping and Pouring}}
\begin{figure}[ht!]
    \begin{subfigure}[b]{0.24\textwidth}
        \includegraphics[width=\textwidth]{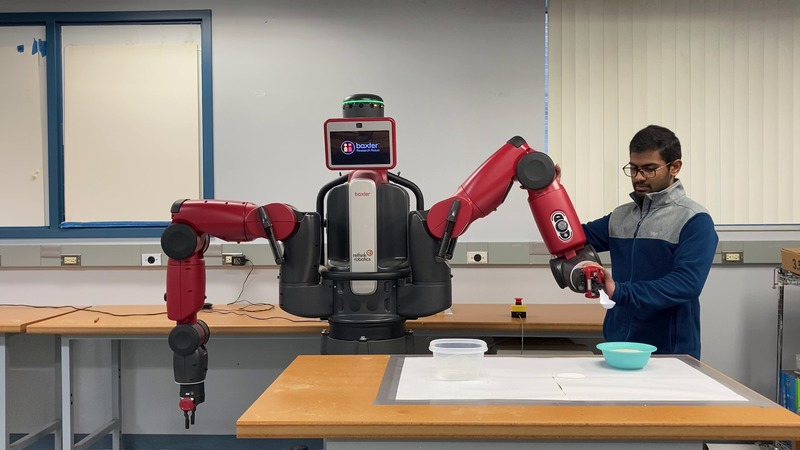}
    \end{subfigure}
    \begin{subfigure}[b]{0.24\textwidth}
        \includegraphics[width=\textwidth]{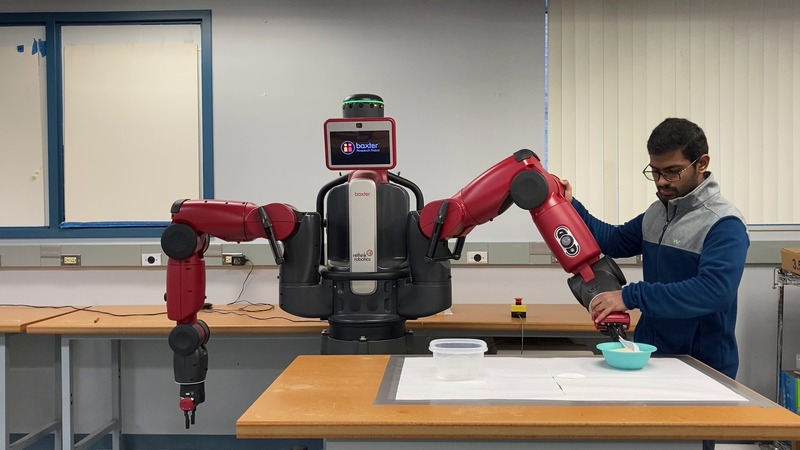}
    \end{subfigure}
    \par\smallskip
    \begin{subfigure}[b]{0.24\textwidth}
        \includegraphics[width=\textwidth]{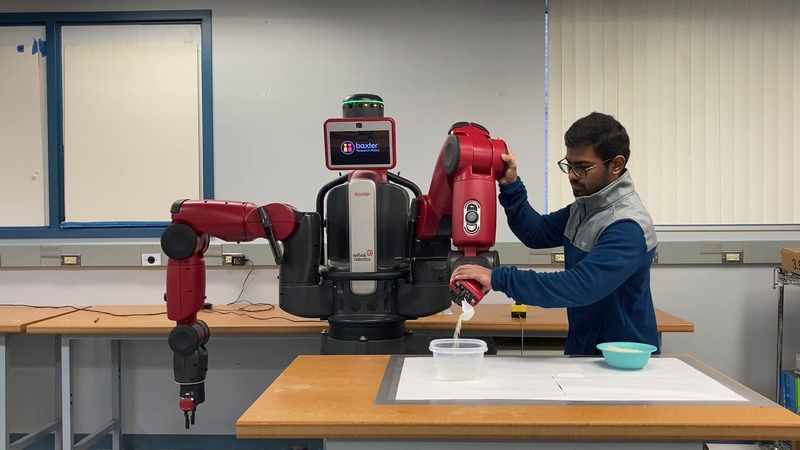}
    \end{subfigure}
    \begin{subfigure}[b]{0.24\textwidth}
        \includegraphics[width=\textwidth]{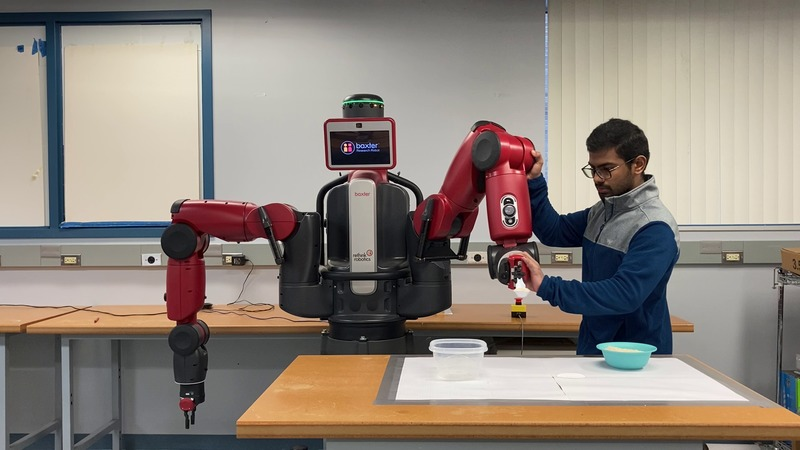}
    \end{subfigure}
    \caption{\textbf{\textsc{Scoop and Pour:}} Demonstration 1\\
    (Clockwise from top left)}
    \label{fig:scoop_and_pour_demonstration}
\end{figure}

In this task the robot scoops rice from one bowl using a spoon which the robot is already grasping and pours it into another bowl which is placed at a different position (see Figure~\ref{fig:scoop_and_pour_demonstration}). The radius of the region of interest used for obtaining the key segments is $20$ cm. %In this experiment, the recorded trajectory was first segmented into a sequence of constant screws. Based on the positions of both the bowls, the key segments of the trajectory which form the task constraints are selected. Using the selected segments the guiding poses are determined. For any given positions of both the bowls, the guiding poses are recomputed and the motion plan is determined. Although the key segments lie near the bowls, due to the nature of ScLERP, the constraint that the orientation of the spoon must be maintained during transfer from one bowl to the other is satisfied without the need for explicitly enforcing it.

Two demonstrations were given with different poses of the objects. For each demonstration we extracted the task-relevant screws, and used it to compute motion plans for eight experimental trials. Thus a total of $16$ experimental trials were performed for this task. Each trial was conducted by varying pose of one or both the bowls. In particular, the heights of the bowls were also changed among some trials, so that merely mimicking the demonstrated path would result in a collision. During each trial, it was ensured that the level of rice in the bowl was the same as in the demonstration. Due to the lack of perception, our framework cannot account for the level of rice in the bowl. All of the trials resulted in successful scooping and transfer of rice from one bowl to another.
%For all the trials conducted, the constraints required for scooping the rice, transferring it to the second bowl and pouring the rice were met.

%\textcolor{red}{For this task, we consider a trial to be a success if the robot can scoop rice from one bowl and pour it into the second bowl while ensuring that the spoon is oriented properly during transfer.}

\begin{figure}[ht!]
    \begin{subfigure}[b]{0.24\textwidth}
        \includegraphics[width=\textwidth]{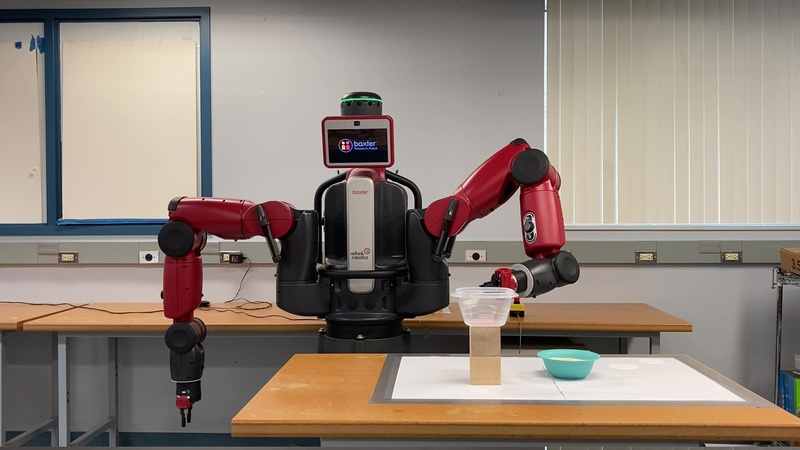}
    \end{subfigure}
    \begin{subfigure}[b]{0.24\textwidth}
        \includegraphics[width=\textwidth]{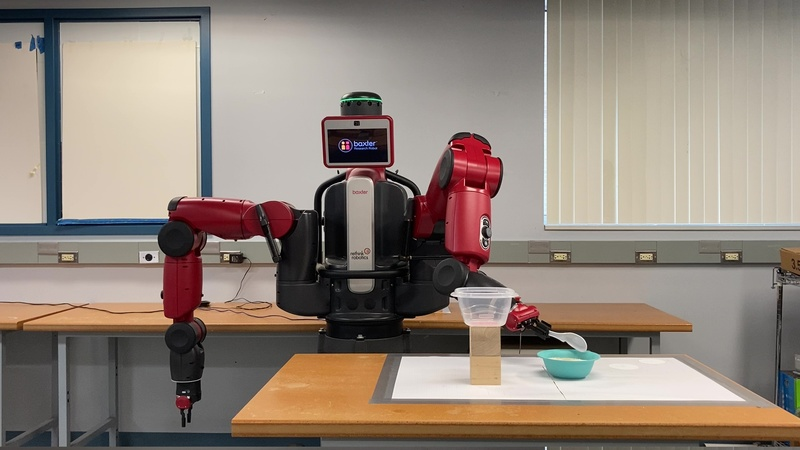}
    \end{subfigure}
    \par\smallskip
    \begin{subfigure}[b]{0.24\textwidth}
        \includegraphics[width=\textwidth]{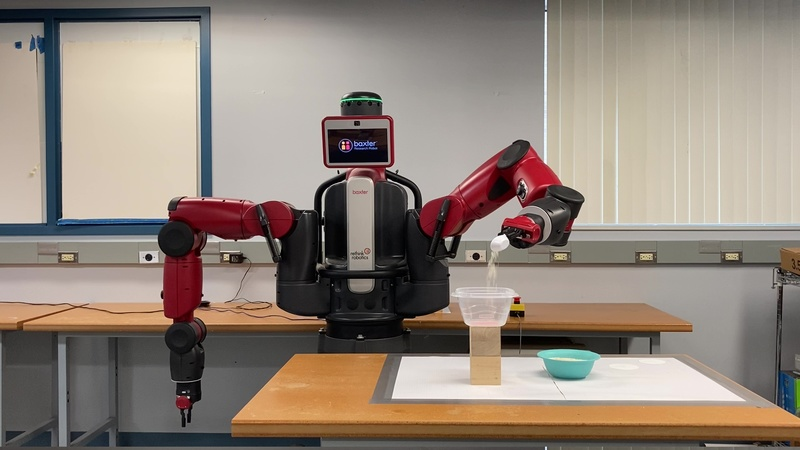}
    \end{subfigure}
    \begin{subfigure}[b]{0.24\textwidth}
        \includegraphics[width=\textwidth]{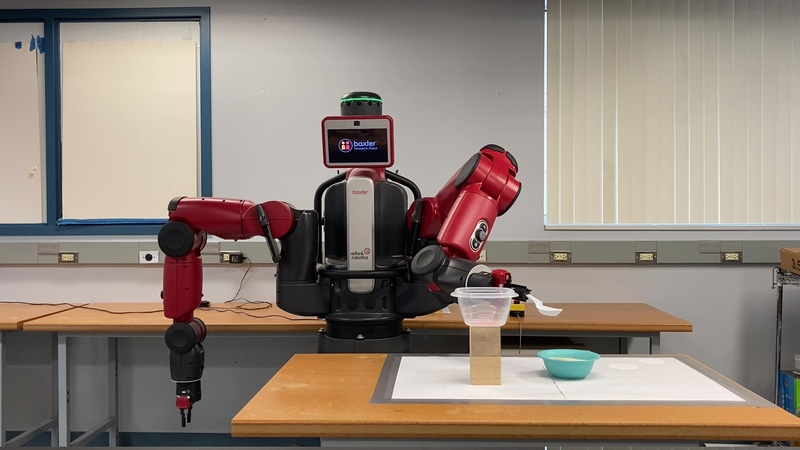}
    \end{subfigure}
    \caption{\textbf{\textsc{Scoop and Pour:}} Trial 8 using Demonstration 1\\
    (Clockwise from top left)}
    \label{fig:scoop_and_pour_demo_1_trial_8}
\end{figure}
%\textcolor{blue}{Again, the table is not informative. We can just say in words in the text that we performed two sets of experiments with two demonstrations and for each demonstration we looked at $8$ generalization cases.}

\subsubsection{\textbf{Arranging Dishes}}
In this task we pick a plate placed on the table and insert it into one of the slots of a dish rack that is placed on the table. This exemplar task was chosen because it consists of a challenging collision avoidance problem when the dish has to be placed in the dish rack. The geometry of the dishes and the slots of the dish rack are such that putting in the dishes top down vertically would result in the dish getting jammed and not reaching the bottom of the rack. The dish has to be put in at an angle and moved in a way that is hard to describe in words or write equations for, but easier for a human to demonstrate. There were $3$ different demonstrations provided and for each demonstration, we performed $8$ experimental trials. Thus, a total of $24$ experimental trials were performed. To extract the key segments the region of interest were chosen to be a sphere of radius $15$ cm for the dish and a cube of side $45$ cm for the dish rack. The key segments were extracted from the demonstration and the guiding poses were determined with the knowledge of the new pose of the task related objects (plate and dish rack). The motion plan was determined using the guiding poses. Also, the grasping information, i.e., where the gripper should be closed relative to the pose of the object is determined from the demonstration and is used to grasp the object during execution.
%We performed three iterations for this experiment. Each iteration consisted of provided a single demonstration of picking up a plate and placing it in the rack. For each iteration of the experiment, we performed multiple trials as listed in Table $\ref{arranging_dishes_trials}$. For this task we also performed one trial where using the demonstration we performed the task of arranging multiple dishes in the dish rack successfully (Figure \ref{fig:arranging_dishes_multiple_dishes_trial}).

\begin{table}[h!]
\centering
\begin{tabular}{||c|c|c|c||} 
    \hline
    \hline
    \multirow{2}{*}{\textbf{Demonstration}} & 
    \textbf{Number of} & 
    \multirow{2}{*}{\textbf{Successful Trials}} & 
    \multirow{2}{*}{\textbf{Failed Trials}}\\
    & \textbf{Trials} & & \\
    \hline
    \hline
    Demonstration 1 & 8 & 7 & 1\\
    \hline
    Demonstration 2 & 8 & 8 & 0\\
    \hline
    %Demonstration 3 & 12 & 9 & 3\\
    Demonstration 3 & 8 & 5 & 3\\
    \hline
    \hline
\end{tabular}
\caption{Arranging Dishes - Trials}
\label{arranging_dishes_trials}
\end{table}

For this task, a trial is considered to be a success if the robot can pick up the dish from the table and insert it into the correct slot in the dish rack. If not, the trial is considered to be a failure. Table~$\ref{arranging_dishes_trials}$ shows the results of the eight trials for all the three demonstrations. Among the different trials, the pose of the dish, the dish rack, as well as the slot in which the dish is being inserted was varied.
All the failures are related to the joint accuracy limits of Baxter. Due to these inaccuracies, the plate was either placed in the neighbouring slot instead of the intended slot or hit the raised edges of the rack and got stuck.

\begin{figure}[h!]
    \begin{subfigure}[b]{0.159\textwidth}
        \includegraphics[width=\textwidth]{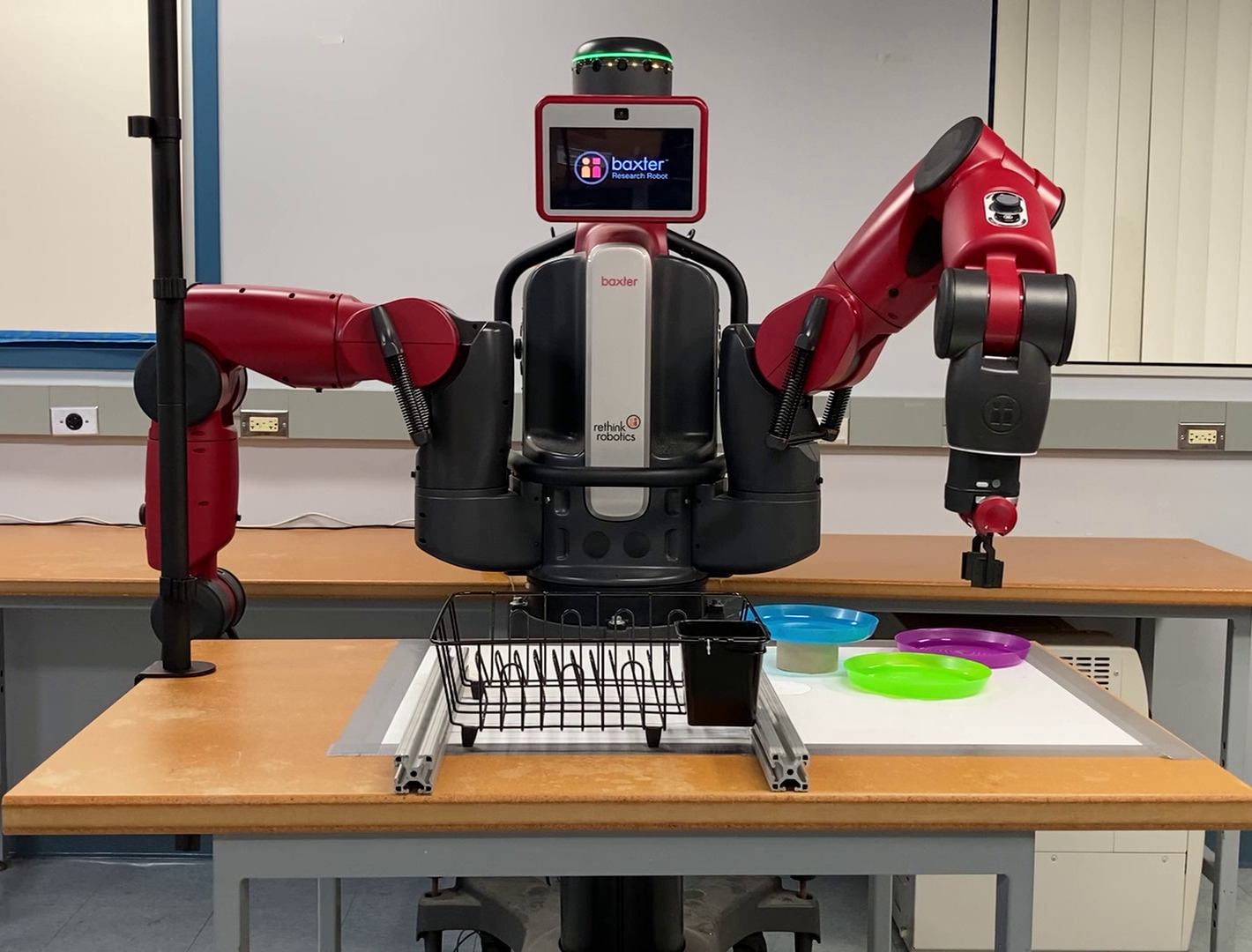}
    \end{subfigure}
    \begin{subfigure}[b]{0.159\textwidth}
        \includegraphics[width=\textwidth]{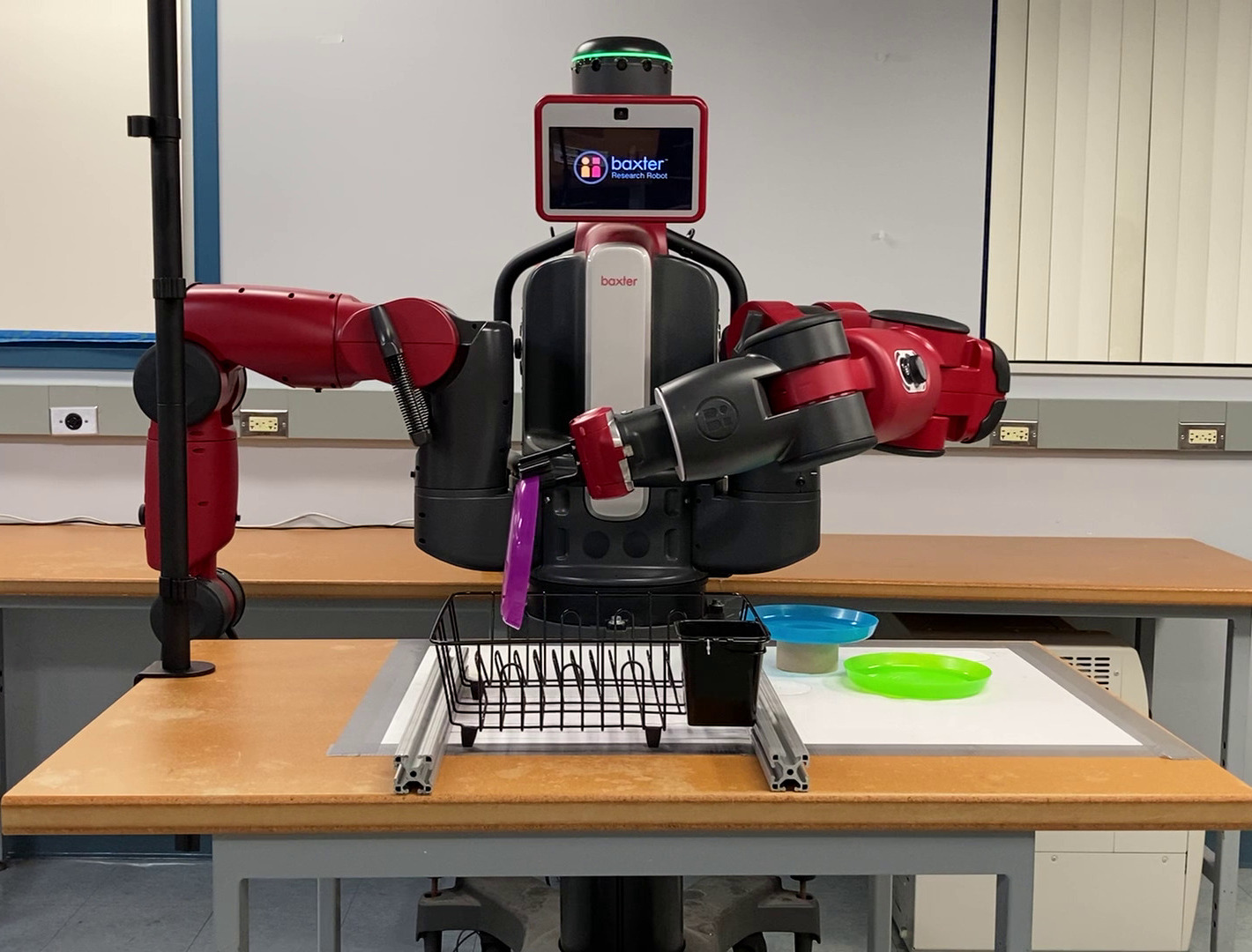}
    \end{subfigure}
    \begin{subfigure}[b]{0.159\textwidth}
        \includegraphics[width=\textwidth]{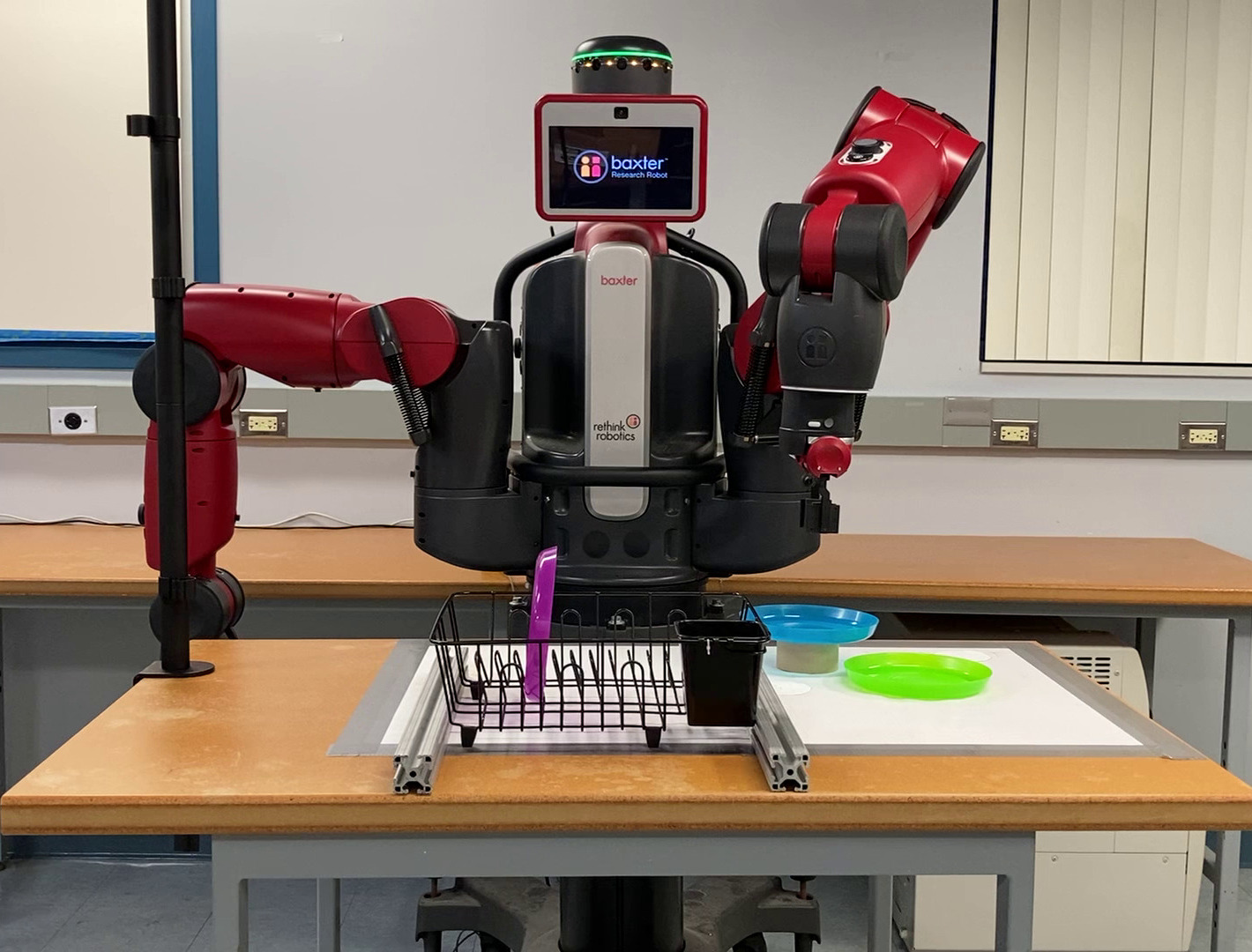}
    \end{subfigure}
    \par\smallskip
    \begin{subfigure}[b]{0.159\textwidth}
        \includegraphics[width=\textwidth]{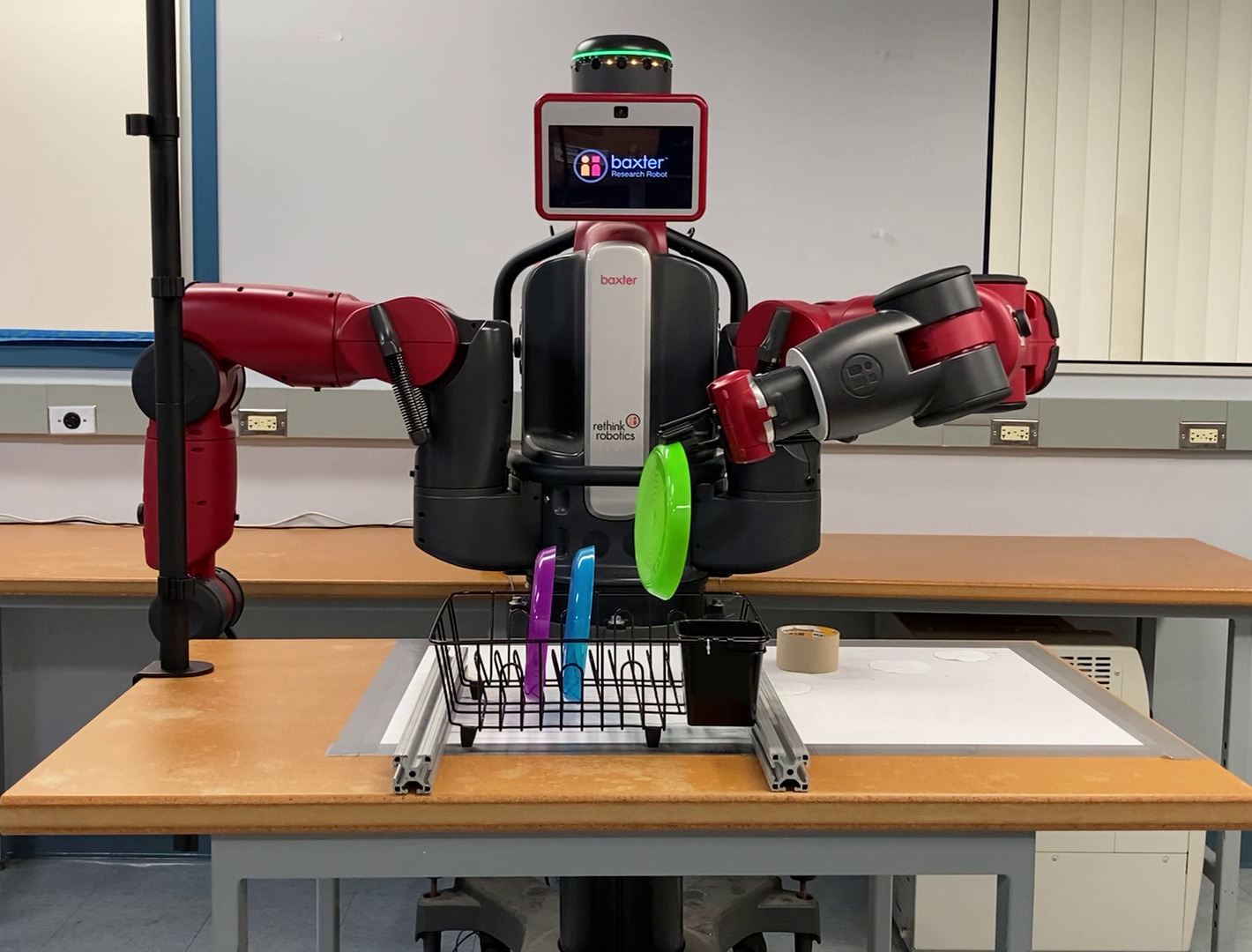}
    \end{subfigure}
    \begin{subfigure}[b]{0.159\textwidth}
        \includegraphics[width=\textwidth]{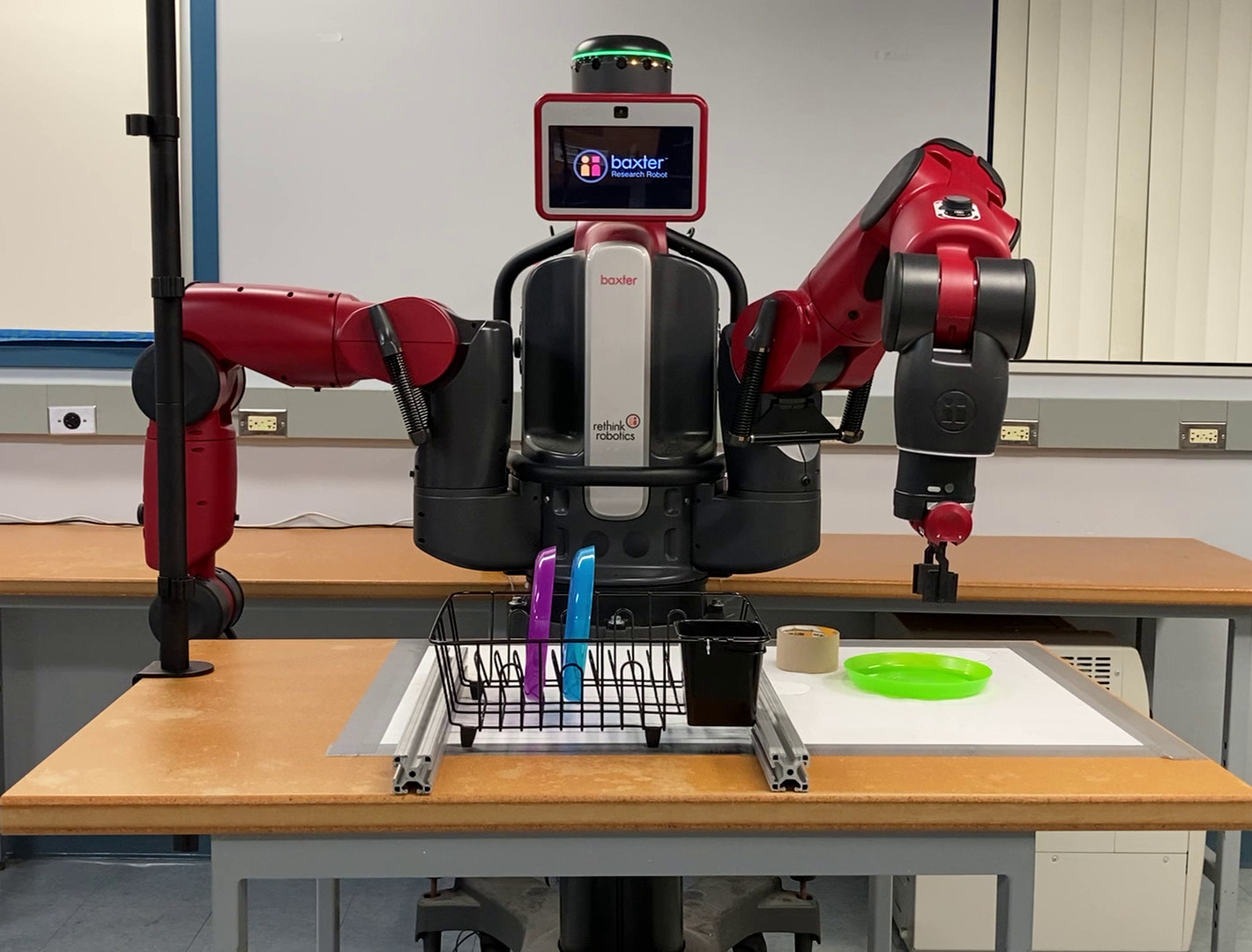}
    \end{subfigure}
    \begin{subfigure}[b]{0.159\textwidth}
        \includegraphics[width=\textwidth]{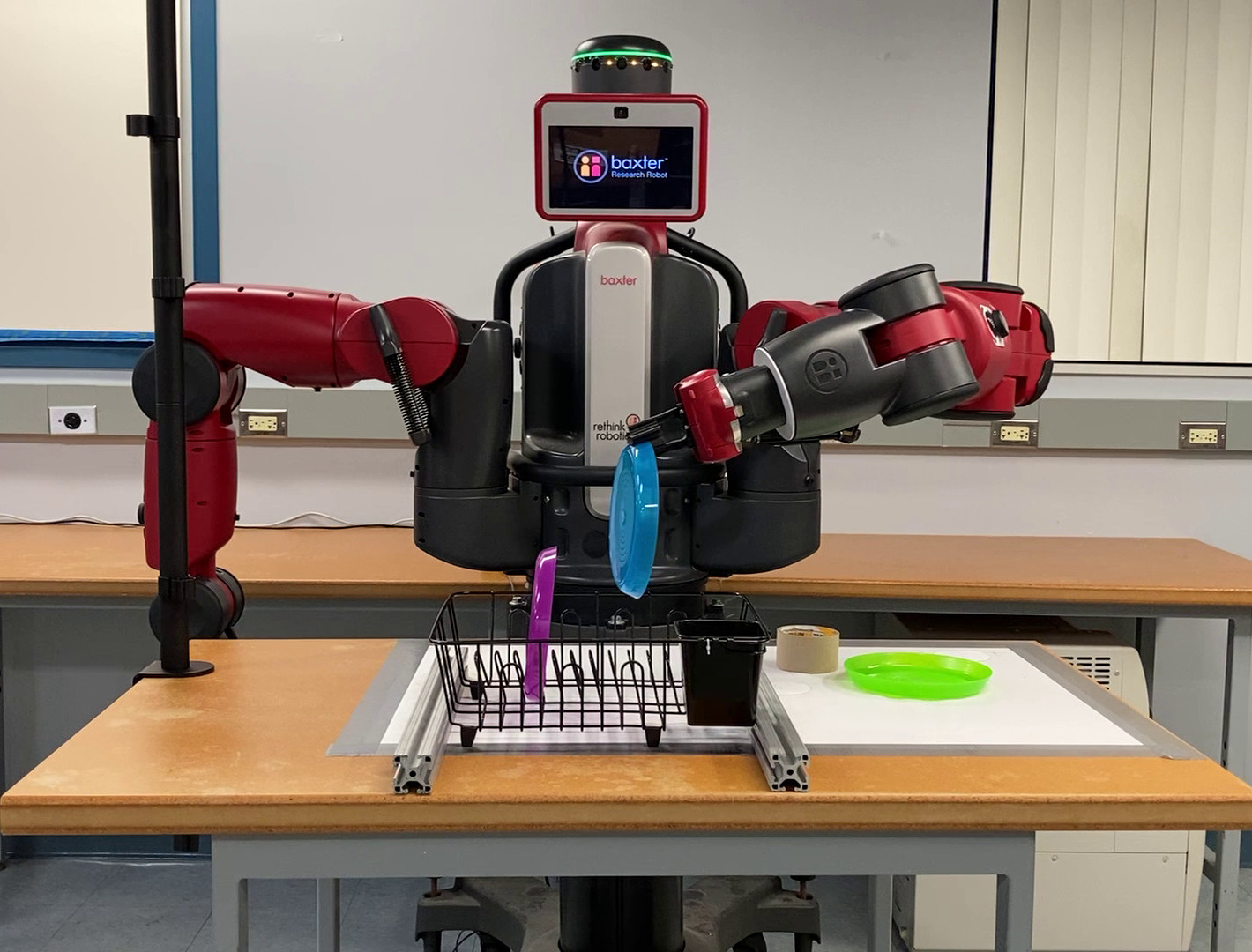}
    \end{subfigure}
    \caption{\textbf{\textsc{Arranging Dishes:}} Arranging multiple dishes using a single demonstration (Clockwise from top left)}
    \label{fig:arranging_dishes_multiple_dishes_trial}
\end{figure}

We also performed one trial of inserting multiple dishes into the dish rack using the example of insertion of a single dish. Figure~\ref{fig:arranging_dishes_multiple_dishes_trial} shows snapshots from the experimental trial. This trial was meant as a proof-of-concept demonstration and we plan to perform more extensive experimental evaluation of this scenario in future work.

\begin{figure}[h!]
    \begin{subfigure}[b]{0.24\textwidth}
        \includegraphics[width=\textwidth]{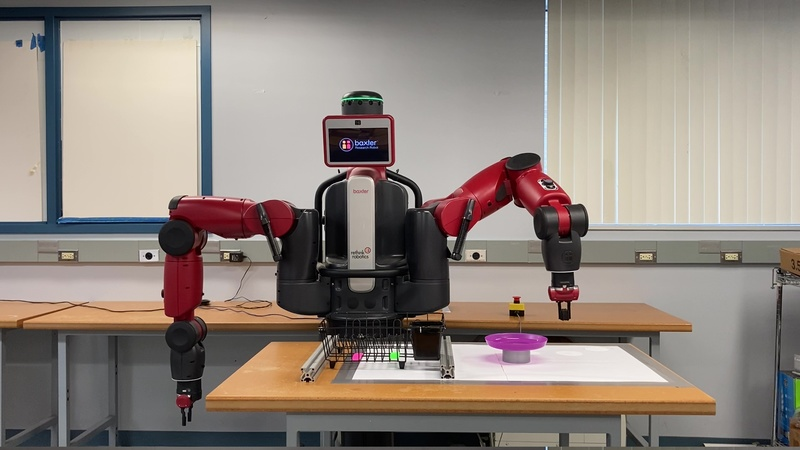}
    \end{subfigure}
    \begin{subfigure}[b]{0.24\textwidth}
        \includegraphics[width=\textwidth]{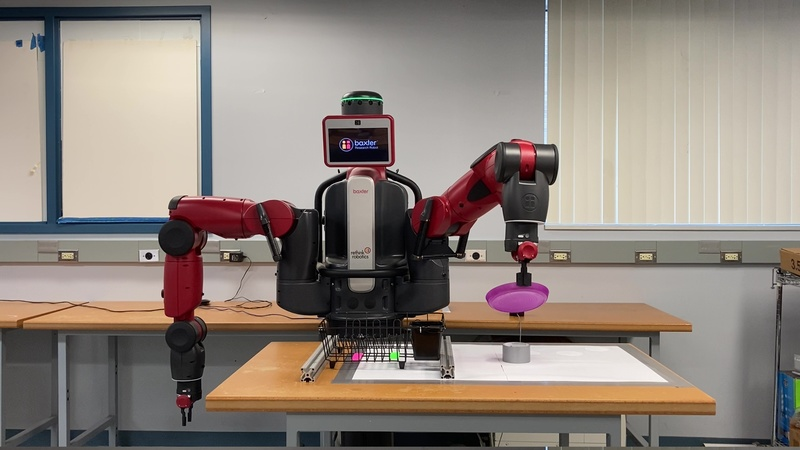}
    \end{subfigure}
    \par\smallskip
    \begin{subfigure}[b]{0.24\textwidth}
        \includegraphics[width=\textwidth]{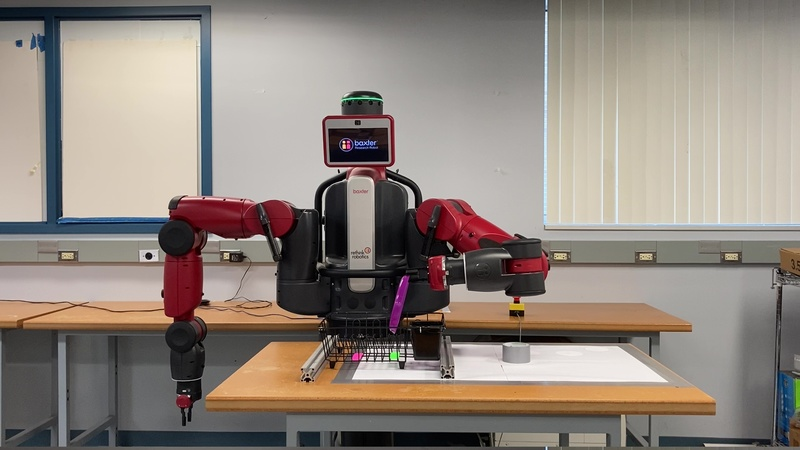}
    \end{subfigure}
    \begin{subfigure}[b]{0.24\textwidth}
        \includegraphics[width=\textwidth]{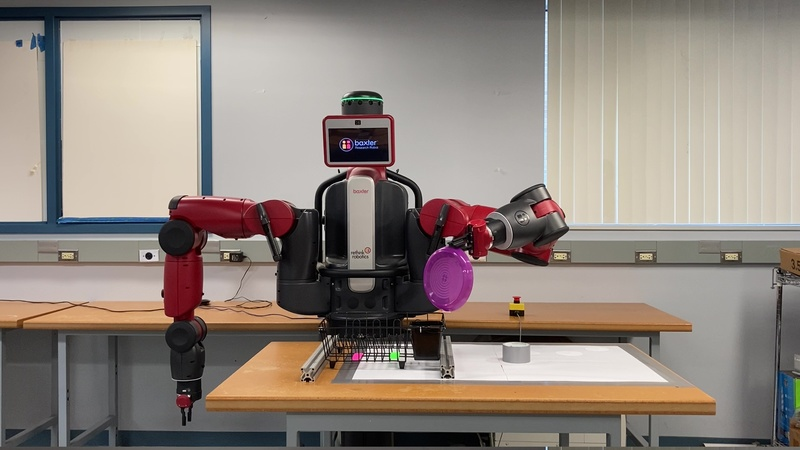}
    \end{subfigure}
    \caption{\textbf{\textsc{Arranging Dishes:}} Trial 12 using Demonstration 3\\(Clockwise from top left)}
    \label{fig:arranging_dishes_demo_3_trial_12}
\end{figure}

\subsection{Discussion}
Since we are trying to extract the implicit constraints embedded in a motion through a kinesthetic guidance demonstration, there is an underlying assumption that the demonstrator is benevolent or trying to help the robot. As discussed above, although there is no strict requirement about how an object should be held during the task, it is assumed that the object is held in such a way that the required motion can be performed and there is no slippage between the object and the robot gripper during the demonstration. 

In principle, the extracted screw segments can be used to transfer the constraints to any task instance and provide us a feasible motion plan in the task space. However, within the ScLERP based motion planner, we are using the Jacobian pseudoinverse to compute the joint space motion from the task space motion. Thus, our motion planner is susceptible to the limitations of the Jacobian pseudoinverse based planning schemes, like hitting joint limits, especially if the robot is working near the boundary of its workspace. Although there are solutions for these problems that utilize the nullspace of the Jacobian to move away from the joint limits \cite{liegeois1997}, these solutions do not guarantee that the planner will not get stuck. For mobile manipulation, one possible solution is to position the base in such a way that the task region is not too close to the boundary of the workspace (which we have done in our experiments). Another possible solution is to evaluate its current demonstration for different task instances in simulation and then actively seek demonstration from a human in regions that it gets stuck. We are currently exploring this approach.  

\section{Conclusion and Future Work}

The expression of task constraints as a sequence of constant screws allows us to exploit the structure of motion present in complex manipulation tasks. While expressing these task constraints in the joint space is a hard problem, expressing them as a sequence of constant screws allows us to represent them in a coordinate-invariant manner. In this paper, we have presented an approach to use a single kinesthetic demonstration for extracting the task constraints as a sequence of constant screws. Using this approach, we also show how these computed constant screws can be used to plan motion for a new task instance. For evaluating this approach, we conducted multiple experiments to manipulate articulated objects and perform complex manipulation tasks from a single demonstration. For articulated objects, we were able to generalize to variations in position and orientation of where the robot grasps the object for manipulation. For complex manipulation tasks, we were able to generalize to variations in position and orientation of the task related objects.

The expression of task constraints as a sequence of constant screws in the task-space would potentially allow us to provide demonstrations on one robot and execute the task on another robot with completely different hardware architectures. The proposed method does not take collision avoidance into consideration. Either, there might be an obstacle on the interpolated path between the guiding poses that the end-effector needs to follow, or there might be an obstacle that might come in contact with one of the robot links even when the end-effector does not collide with objects when following the required interpolated path. While the latter case can be overcome by integrating obstacle avoidance in the ScLERP based motion planner as proposed in \cite{sclerp_lcp}, the former case requires modifying the task constraints. In future work we plan on incorporating collision avoidance with the current approach. Also, the heuristic used to determine the guiding poses associated with each object can be improved.

\nocite{Murray1994}
\nocite{DeSchutter2009}
\nocite{Vochten2019}
\nocite{Mike-RSS-13}
\nocite{Mohammadi2021}

\bibliographystyle{abbrv}
\bibliography{references}

\end{document}